\pdfoutput=1
\documentclass[10pt,twocolumn,letterpaper]{article}
\usepackage{preprintstyle}

\usepackage{cuted}   %
\usepackage{caption} %

\usepackage{multirow}
\usepackage{makecell}
\usepackage{colortbl}       %
\usepackage{pifont}         %
\usepackage{bm}
\usepackage{mathtools}
\usepackage{amsthm}
\usepackage{xspace}
\usepackage{algorithm}
\usepackage{algpseudocode}
\usepackage{tabularx}

\theoremstyle{plain}
\newtheorem{lemma}{Lemma}
\newtheorem{proposition}{Proposition}
\newtheorem{corollary}{Corollary}
\theoremstyle{definition}

\theoremstyle{remark}
\newtheorem{remark}{Remark}
\AtEndPreamble{%
  \crefname{lemma}{Lemma}{Lemmas}\Crefname{lemma}{Lemma}{Lemmas}%
  \crefname{proposition}{Prop.}{Props.}\Crefname{proposition}{Proposition}{Propositions}%
  \crefname{corollary}{Cor.}{Cors.}\Crefname{corollary}{Corollary}{Corollaries}%
  \crefname{definition}{Def.}{Defs.}\Crefname{definition}{Definition}{Definitions}%
  \crefname{algorithm}{Alg.}{Algs.}\Crefname{algorithm}{Algorithm}{Algorithms}%
  \crefname{equation}{Eq.}{Eqs.}\Crefname{equation}{Equation}{Equations}%
}

\definecolor{cK1}{HTML}{E87BA4}        %
\definecolor{cK2}{HTML}{B8461B}        %
\definecolor{cK4}{HTML}{184F95}        %
\definecolor{cTeacher}{HTML}{0B0B0B}   %
\definecolor{cRealGRN}{HTML}{1BAF7A}   %
\definecolor{cPaired}{HTML}{7B3FA0}    %
\definecolor{cNative}{HTML}{C9A227}    %
\definecolor{cAbl}{HTML}{8A8985}       %
\definecolor{oursrow}{HTML}{EEF3FA}    %

\definecolor{cWin}{HTML}{5A35C8}       %
\definecolor{cRS}{HTML}{C2297A}        %
\definecolor{cSB}{HTML}{0B9358}        %
\definecolor{cEH}{HTML}{1B9E9E}        %
\definecolor{cCK}{HTML}{37474F}        %

\providecommand{\Kset}{\mathcal{K}}
\providecommand{\tagW}{\textsf{W}}
\providecommand{\tagRS}{\textsf{RS}}
\providecommand{\tagSB}{\textsf{OS}}   %
\providecommand{\tagEH}{\textsf{NS}}   %
\providecommand{\tagCK}{\textsf{RCK}}   %

\providecommand{\PEND}[2][]{\textcolor{red}{\textbf{[\if\relax\detokenize{#1}\relax\texttt{\detokenize{#2}}\else#1\fi]}}}

\newcommand{\method}{\mbox{FastGRN}\xspace}  %

\DeclareMathOperator{\softmax}{softmax}
\newcommand{\E}{\mathbb{E}}
\newcommand{\KL}{D_{\mathrm{KL}}}
\newcommand{\TC}{\mathrm{TC}}                %
\newcommand{\sg}{\operatorname{sg}}          %
\newcommand{\Vsz}{V}                         %
\newcommand{\qmin}{q_{\min}}
\newcommand{\qmax}{q_{\max}}
\newcommand{\kq}{k_q}                        %
\newcommand{\vx}{\bm{x}}
\newcommand{\vz}{\bm{z}}
\newcommand{\veps}{\bm{\epsilon}}
\newcommand{\vc}{\bm{c}}
\newcommand{\Gs}{G_\theta}                   %
\newcommand{\Fs}{F_\psi}                     %
\newcommand{\Rs}{R}                          %

\newcommand{\FigPH}[3]{%
  {\setlength{\fboxsep}{0pt}\fbox{%
    \begin{minipage}[c][#2][c]{\dimexpr#1-2\fboxrule\relax}\centering\footnotesize#3\end{minipage}}}}
\newcommand{\IncludeOrPH}[5][]{%
  \IfFileExists{#2}{\includegraphics[#1]{#2}}{\FigPH{#3}{#4}{#5}}}
\newcommand{\InputOrPH}[2]{\IfFileExists{#1}{\input{#1}}{#2}}
\newcommand{\TablePH}[3]{\begin{table}[t]\centering\FigPH{\linewidth}{#3}{#2}\caption{#2}\label{#1}\end{table}}
\newcommand{\TablePHwide}[3]{\begin{table*}[t]\centering\FigPH{\linewidth}{#3}{#2}\caption{#2}\label{#1}\end{table*}}

\IfFileExists{tables/tab_macros.tex}{
\providecommand{\na}{--}
\providecommand{\regrn}{FastGRN}

\makeatletter
\@ifundefinedcolor{cWin}{\definecolor{cWin}{HTML}{5A35C8}}{}
\@ifundefinedcolor{cRS}{\definecolor{cRS}{HTML}{C2297A}}{}
\@ifundefinedcolor{cSB}{\definecolor{cSB}{HTML}{0B9358}}{}
\@ifundefinedcolor{cEH}{\definecolor{cEH}{HTML}{1B9E9E}}{}
\@ifundefinedcolor{cCK}{\definecolor{cCK}{HTML}{37474F}}{}
\makeatother
\providecommand{\tagW}{\textsf{W}}
\providecommand{\tagRS}{\textsf{RS}}
\providecommand{\tagSB}{\textsf{OS}}
\providecommand{\tagEH}{\textsf{NS}}
\providecommand{\tagCK}{\textsf{RCK}}   %

\providecommand{\PEND}[2][]{\textcolor{red}{\textbf{[\if\relax\detokenize{#1}\relax\texttt{\detokenize{#2}}\else#1\fi]}}}
\providecommand{\Kset}{\mathcal{K}}
}{}

\usepackage[pagebackref,breaklinks,colorlinks,allcolors=black]{hyperref}
\hypersetup{pdftitle={FastGRN: Few-Step Generative Refinement Networks},
  pdfauthor={Haoyang Hu, Masataka Seo, Liang Li, Yen-wei Chen}}

\title{\method: Few-Step Generative Refinement Networks}

\author{Haoyang Hu$^{1}$\qquad Masataka Seo$^{2}$\qquad Liang Li$^{1}$\qquad Yen-wei Chen$^{1}$\\
{\normalsize $^{1}$Ritsumeikan University\qquad $^{2}$Osaka Institute of Technology}}

\begin{document}
\maketitle

\IfFileExists{figs/teaser.pdf}{%
\begin{strip}
\centering
\vspace{-1.2em}
\includegraphics[width=\linewidth]{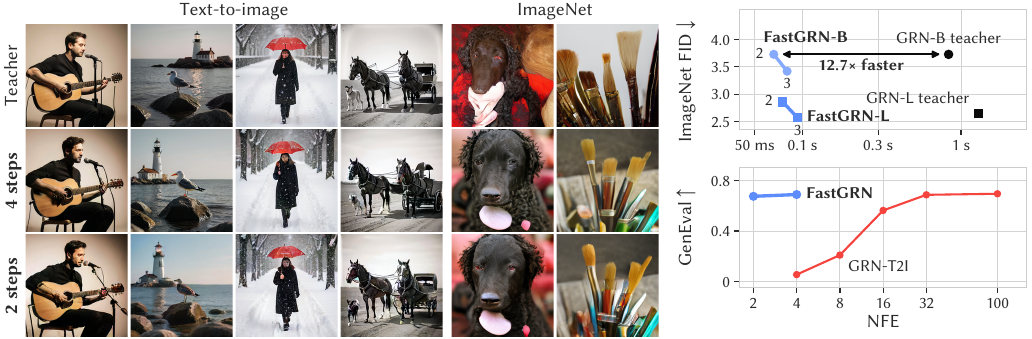}
\captionsetup{hypcap=false}
\vspace{-1.6em}
\captionof{figure}{\textbf{\method turns a 50-step Generative Refinement Network into a 2--4 step generator.}
Left: $256^2$ samples (GRN-T2I 2B, same noise per column; GRN-B). Right: ImageNet FID-50k \vs latency
(numerals: steps; \cref{tab:c2i}) and GenEval \vs NFE against the teacher's own sampler with fewer steps.}
\captionsetup{hypcap=true}
\label{fig:teaser}
\vspace{0.3em}   %
\end{strip}
}{}

\begin{abstract}
Generative Refinement Networks (GRN) generate an image by repeatedly re-predicting and refining every token of a
discrete latent. They reach high fidelity, but need dozens of Markov steps. We therefore train few-step GRNs on the
distribution of their own samples. We use GRN's own uniform-state corruption to perform distribution matching on
discrete tokens, which lets a discrete generator learn by back-propagation, and we pair noise with samples in the
manner of rectified flow. We find that the steps of a few-step GRN do not carry the same amount of information or the
same repair load. We therefore exploit GRN's discrete refinement during training and build few-step GRNs in three parts.
\emph{Oversampling} samples the
student with more steps than it was trained for and refines its judge on these samples. \emph{Non-uniform sampling}
trains only the last step and hands it most of the information at sampling time. \emph{Refinement confidence keeping} holds on to
the tokens the student is already sure of. We train a fast class-conditional generator on ImageNet $256^2$ and, with
the same method, a 2B text-to-image model. On ImageNet $256^2$, the 3-step \method beats the FID of GRN. On a 2B binary-token text-to-image GRN, a 4-step
student also nearly matches the GenEval of GRN-T2I 2B.
\end{abstract}

\section{Introduction}
\label{sec:intro}

Discrete-token image generators have advanced quickly, from masked~\cite{chang2022maskgit,chang2023muse} to
autoregressive and scale-wise models~\cite{esser2021vqgan,sun2024llamagen,tian2024var,han2025infinity}. Generative
Refinement Networks (GRN)~\cite{han2026grn} are the newest member of this family. GRN encodes an image into a
near-lossless map of discrete tokens with a hierarchical binary quantizer (HBQ). At every step, a transformer
re-predicts \emph{all} tokens from a partly corrupted copy of its previous prediction. This design gives strong
results on ImageNet $256^2$ and GenEval, at the price of 50 refinement steps with classifier-free guidance
(CFG)~\cite{ho2021cfg}.

Simply cutting the number of steps does not work. With 4 of the 50 steps, GRN's ImageNet samples fall
apart (Fig.~\ref{fig:nfe}). Every step revises every token, so we might expect later steps to repair a rushed
start, yet in practice they cannot.

The problem is GRN's factorized output head. GRN corrupts a token by keeping it with some probability and
otherwise replacing it with a uniformly random one, which matches the marginal of uniform-state discrete
diffusion~\cite{austin2021d3pm,schiff2025udlm,sahoo2025duo}. Seen this way, the few-step error is the
conditional total correlation~\cite{watanabe1960tc} of the tokens given the current state (Prop.~\ref{prop:tc}).
A factorized predictor trained with cross-entropy cannot represent this dependence. So a few-step GRN has to
learn the joint law of its own samples, which is what distribution matching~\cite{yin2024dmd,yin2024dmd2} does for
continuous generators.

\paragraph{Discrete distribution matching.}
Distribution matching minimizes a reverse KL between the generator's distribution and the data distribution. A frozen
real score and an online fake score judge the samples, and the loss is back-propagated through the generator's
samples by reparameterization. Tokens cannot be reparameterized, so existing discrete distillation methods fall back
on policy gradients~\cite{zheng2026didi}, relaxed tokens~\cite{zhu2026softdimo}, straight-through
estimators~\cite{wang2026fpd}, approximate Jacobians~\cite{zhu2025dimo} or
discriminators~\cite{zheng2026didi,zhu2026softdimo}. Our core observation is that GRN is close to discrete diffusion.
GRN's uniform-state corruption makes all of these unnecessary. The gradient of the reverse KL between the corrupted student and data laws
with respect to each token's logits follows directly from concrete scores~\cite{meng2022concrete,lou2024sedd}, and the
posteriors of the real and fake scores give these concrete scores. For each token, the gradient simply
raises the logits of the values that the real score finds more plausible than the fake score, and lowers the others.

\begin{figure*}[t]
  \centering
  \IncludeOrPH[width=\linewidth]{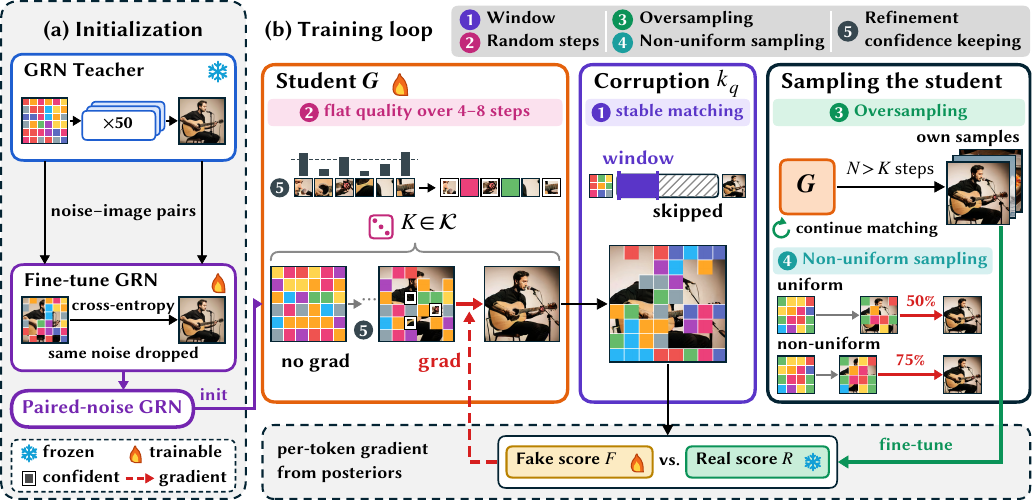}{\linewidth}{6.0cm}{\textbf{Framework placeholder}}
  \vspace{-1.5em}
  \caption{\textbf{\method trains a few-step GRN on its own samples.} \textbf{(a)}~Paired-noise initialization.
  \textbf{(b)}~Only the student's last step receives the gradient of \cref{eq:grad}.
  \textbf{\ding{182}}~The window (\tagW) skips the nearly clean noise levels. \textbf{\ding{183}}~Random steps (\tagRS) redraw
  $K$ every iteration. \textbf{\ding{184}}~Oversampling (\tagSB) fine-tunes the real score on $N>K$-step samples.
  \textbf{\ding{185}}~Non-uniform sampling (\tagEH) enlarges the trained step's share of the path (red).
  \textbf{\ding{186}}~Refinement confidence keeping (\tagCK) keeps confident tokens.}
  \label{fig:framework}
\end{figure*}

\paragraph{Training and sampling.}
\method (\cref{fig:framework}) builds on this gradient in three parts. The first part trains the student,
starting from a paired-noise initialization~\cite{redi2025}. \textbf{(i)~A window} restricts
matching to moderate noise levels. Without the window,
few-step matching collapses into blank images. \textbf{(ii)~Random steps} draw a new number of student steps at
every iteration. One shared fake score matches them all (Prop.~\ref{prop:shared}). A single student then
serves 4 to 8 steps at flat quality, while a fixed-step student blurs below its budget and drifts above it. The
second part changes how the student is sampled (\cref{sec:sampling}), because the steps of a few-step GRN carry
different amounts of information and different repair loads. \textbf{(iii)~Oversampling} samples the
$K$-step student with $N>K$ steps, more than it was trained for, and fine-tunes its real-score judge on these samples. This way, the student
learns to do in $K$ steps what used to take $N$.
\textbf{(iv)~Non-uniform sampling} redistributes the information that each sampling step of a few-step student
carries. In the 2-step student, for example, the second step does 75\% of the refinement, and the 2-step student
reaches the quality of the 4-step one. The third
part, \textbf{(v)~refinement confidence keeping} (\cref{sec:conf}), holds on to the tokens the student is already
sure of.

With non-uniform sampling on top of oversampling, the 2-step \method-B student matches the FID of its teacher, which needs 79 network
function evaluations (NFE), and runs 12.7$\times$ faster per image (\cref{fig:teaser}). It is the only GAN-free student of a discrete generator
that matches its teacher with two network calls (\cref{tab:compare}). On text-to-image, even our 2-step student scores a
higher GenEval than the few-step students of other discrete generators (\cref{tab:t2i_compare}).

\noindent\textbf{Contributions.}
\begin{itemize}[leftmargin=*,itemsep=1pt,topsep=2pt]
  \item \textbf{Discrete distribution matching.} The per-token reverse-KL gradient under uniform-state
  corruption is computed directly from two posterior networks.
  \item \textbf{Three parts that turn the resulting gradient into a few-step GRN}, namely the window with random steps, oversampling with
  non-uniform sampling, and refinement confidence keeping.
  \item \textbf{We propose the first few-step GRN} models, covering ImageNet and a 2B text-to-image model.
\end{itemize}

\section{Related work}
\label{sec:related}

\paragraph{Discrete visual generation.}
Masked~\cite{chang2022maskgit,chang2023muse,lezama2022tokencritic,bai2025meissonic,xie2025showo} and autoregressive
or scale-wise~\cite{esser2021vqgan,sun2024llamagen,tian2024var,han2025infinity} token generators compete with continuous
diffusion and flow models~\cite{ho2020ddpm,song2021sde,lipman2023flow,rombach2022ldm,peebles2023dit,esser2024sd3,li2024mar,li2025jit}.
Unlike masked models, GRN~\cite{han2026grn} re-predicts every token at every step. Its corruption is the uniform-state
kernel of discrete diffusion~\cite{austin2021d3pm,campbell2022ctmc}, which UDLM~\cite{schiff2025udlm},
Duo~\cite{sahoo2025duo}, discrete flow matching~\cite{campbell2024dfm,gat2024dfm} and URSA~\cite{deng2025ursa} also
use. GRN does not draw this connection, and our gradient is built on it.

\paragraph{Few-step continuous generation.}
Steps have been reduced by progressive distillation~\cite{salimans2022progressive}, consistency
models~\cite{song2023consistency,luo2023lcm,lu2025scm}, rectified flow~\cite{liu2023rectified,liu2024instaflow} and
flow maps~\cite{frans2025shortcut,geng2025meanflow}. Score and distribution matching~\cite{luo2023diffinstruct,wang2023prolificdreamer,yin2024dmd,yin2024dmd2,xu2025fdistill,wu2026dpdmd}
trains a generator through a frozen real score and an online fake score, sometimes together with adversarial
losses~\cite{sauer2024add}. Our windowed matching is a discrete counterpart of this reverse KL. Comparing distributions after a noise
kernel goes back to spread divergences~\cite{zhang2020spread} and the reverse diffusive KL~\cite{he2025rdkl}.
Reflow~\cite{liu2023rectified} re-trains a model on teacher noise--sample pairs, and we use such pairs as an
initialization.

\paragraph{Few-step discrete generation.}
Masked image models have been distilled by consistency distillation, as in Show-o Turbo~\cite{xu2025showoturbo}, and
by distribution matching. Di[M]O~\cite{zhu2025dimo} matches a token-level Jeffrey divergence with an auxiliary model
and replaces the Jacobian at the intermediate state with that of the one-step output. Soft-Di[M]O~\cite{zhu2026softdimo}
relaxes tokens to expected embeddings and adds a GAN loss, and FPD~\cite{wang2026fpd} uses fixed-point matching with
a straight-through estimator. Distilled Decoding~\cite{liu2025distilleddecoding} reduces autoregressive models to one
or two steps. For discrete diffusion, Duo~\cite{sahoo2025duo} distills a uniform-state model by discrete consistency
distillation, Di4C~\cite{hayakawa2025di4c} uses mixture students to model correlations, and SDTT~\cite{deschenaux2025sdtt}
self-distills through time. DiDi-Instruct~\cite{zheng2026didi} matches integral KLs with policy gradients and a
discriminator, D-MMD~\cite{hoogeboom2026dmmd} distills with a discrete moment-matching objective, and Gao and
Sun~\cite{gao2025ecdm} match conditionals recovered from density ratios. ReDi~\cite{redi2025} reduces the
factorization error of discrete flows by rectifying the coupling, and our paired-noise initialization carries this
idea to GRN. Discrete MeanFlow-style generators~\cite{khan2026discretemeanflow,ouyang2026dag} learn interval
transitions, which add nothing beyond the posterior for per-token predictors (Lemma~\ref{lem:redundancy}).

\paragraph{Any-step distillation and self-improvement.}
Shortcut and consistency (trajectory) models~\cite{frans2025shortcut,song2023consistency,kim2024ctm}, flow
maps~\cite{geng2025meanflow,sabour2025ayf} and moment matching~\cite{salimans2024multistep,zhou2025imm} condition a
network on the step size, and distribution matching has multi-step~\cite{yin2024dmd2} and
flexible-step~\cite{luo2025tdm} variants. Random steps need no extra conditioning (Prop.~\ref{prop:shared}).
Oversampling is a form of expert iteration~\cite{anthony2017exit,silver2017alphagozero} and
self-training~\cite{xie2020noisystudent,furlanello2018bornagain}. Prior work retrains the generator itself on its own
samples~\cite{alemohammad2024mad,bertrand2024stability}. In \method, only the judge sees them.

\paragraph{Step grids.}
Sampling schedules are usually optimized for an already trained model~\cite{sabour2024ays,park2025jys,tong2025ld3},
and few-step students, including DMD2~\cite{yin2024dmd2}, train on evenly spaced timesteps. \method treats the
training grid itself as a design choice.

\emph{Positioning.} The discrete distillation methods above estimate, relax or approximate the gradient that
DMD~\cite{yin2024dmd} obtains by reparameterization. Our matching keeps hard tokens and computes the reverse-KL gradient
directly from two posterior networks. It combines the concrete-score--posterior identity of SEDD and UDLM~\cite{lou2024sedd,schiff2025udlm}
with local-expectation gradients~\cite{titsias2015local}. All three parts of \method are new to discrete distribution
matching.

\section{Background: Generative Refinement\texorpdfstring{\protect\newline}{ }Networks}
\label{sec:background}

\paragraph{Tokens and corruption.}
GRN~\cite{han2026grn} encodes an image with the HBQ tokenizer into discrete tokens $\vx$, each taking one of $\Vsz$
values. For class-conditional ImageNet $256^2$, the token map has $16{\times}16$ positions with 16 channels, \ie 4096
tokens with 16 ordered levels. The 2B text-to-image model uses 65{,}536 bits. During training, every token $x^i$ is
corrupted independently at a time $t\in[0,1]$ with the uniform-state kernel
\begin{equation}
  k_t(z^i\mid x^i) \;=\; t\,\delta(z^i,x^i) + \tfrac{1-t}{\Vsz},
  \label{eq:kernel}
\end{equation}
which keeps the token with probability $t$ and otherwise draws it uniformly at random, as in uniform-state
diffusion~\cite{austin2021d3pm,schiff2025udlm}. We write $k_q$ for the same kernel at keep probability $q$. A
transformer with a factorized output head learns to predict the clean tokens from the corrupted tokens $\vz$, the time
and the condition with per-token cross-entropy (CE). At the optimum, it returns the posterior marginals
$p(x^i\mid\vz)$.

\paragraph{Sampling.}
GRN draws a uniform noise map $\veps$ once and refines it on a time grid from 0 to 1. At every step, it samples clean
tokens from the CFG-combined logits at temperature $\tau$. It then keeps each sampled token with probability equal to
the next time on the grid and otherwise re-uses the frozen noise token. GRN-B/L sample with 50 steps and CFG on
$t\in(0.4,1)$ (79 NFE), and the text-to-image model applies CFG at every step (100 NFE). Our students use the same
sampler with $K$ equally spaced steps, and non-uniform sampling moves the first intermediate time $t_1$ to
$0.25$ (\cref{sec:eh}). Since CFG is distilled into the student, it costs $K$ NFE.
$K$ is either fixed, $K=2$ or $4$, or drawn at every iteration from a set $\Kset$ (\cref{sec:randk}), and any
student can also run $N\neq K$ steps.

\section{Method}
\label{sec:method}

We first analyze the few-step error (\cref{sec:why}) and then build \method in three parts
(\cref{sec:train,sec:sampling,sec:conf}), with proofs in the supplement.

\subsection{Why few steps fail}
\label{sec:why}

GRN's output head is factorized. Given the current state, it predicts every token independently. Two tokens that
straddle an object boundary, for example, may each pick a different position for it. Over many steps, the
tokens can still coordinate through the intermediate states, and with few steps there is little room for this. The
best that cross-entropy training can reach among factorized predictors is the product of the posterior marginals.
The error that remains is the conditional total correlation~\cite{watanabe1960tc} of the tokens given the state,
which equals the sum of the per-token entropies minus the joint entropy (Prop.~\ref{prop:tc}).

This is the factorization error studied for discrete diffusion and flows~\cite{redi2025,hayakawa2025di4c}. The
student as a whole, however, need not be factorized. Its last step is factorized only given its latent $\lambda$,
which consists of the noise and the earlier states. The student therefore samples a mixture of product laws, which
can represent dependence between tokens, and distribution matching shapes this mixture directly (\cref{sec:wdm}).
Informative initial noise lowers the error as well (Corollary~\ref{cor:paired}).

\subsection{Training the student by distribution matching}
\label{sec:train}
\paragraph{Paired-noise initialization.}
\label{sec:paired}
GRN is trained with fresh random tokens, so its first step ignores the noise. We first make the noise
informative. We sample noise--image pairs from the guided teacher, keep the frozen initial noise of each sample, and
fine-tune GRN with its own CE objective while filling the dropped tokens with the paired noise
(\cref{fig:framework}a). This way, the first step learns what the noise says about each token
(Corollary~\ref{cor:paired}). The resulting paired-noise GRN initializes the student. It already draws
recognizable objects with two steps, while GRN fine-tuned with independent noise still draws noise
(Tab.~\ref{tab:fewstep}), so the student starts from a much better place.

\paragraph{Windowed distribution matching.}
\label{sec:wdm}
Let $p_{\theta,q}$ and $p_{\mathrm{d},q}$ be the laws of the $K$-step student's samples and of the data (teacher)
after corruption at keep probability $q$. We minimize the reverse KL $\KL(p_{\theta,q}\|p_{\mathrm{d},q})$ averaged
over $q\sim\rho$ (Eq.~(\ref{eq:wdm})), an equal mixture of GRN's training law of $t$ and the uniform law, clamped to
the window $[0.02,\qmax]$. With $\qmax=0.4$, nearly half of the draws land at the upper end, so matching mostly happens at
one moderate noise level. The corruption is invertible at every nonzero keep probability, so each term vanishes only
when the student samples the data law. So the window changes the optimization path and keeps the same minimizer. A
frozen real score (the fine-tuned GRN or the guided teacher) predicts the clean tokens under the data law, and a fake score,
initialized from the fine-tuned GRN, does so under the student's law. The fake score learns online with CE on detached student
samples at all noise levels, with several updates per student update~\cite{heusel2017fid,yin2024dmd2}, and is queried
only inside the window.

\paragraph{Per-token gradient.}
Given its latent $\lambda$ (the noise and the earlier states), the last student step samples token $i$ from
$\pi^i=\softmax(u^i)$, with $u^i$ its tempered logits. The KL gradient needs concrete
scores~\cite{meng2022concrete,lou2024sedd}, the probability of a corrupted state with one token changed over that of
the state itself. Under the uniform kernel, they follow directly from a clean-token prediction
(Eq.~(\ref{eq:delta})), made by the fake score for the student and by the real score for the data.

\begin{proposition}[Discrete distribution matching gradient]
\label{prop:grad}
Let $\Delta^i_a$ be the log concrete score of the fake model minus that of the real model for changing token $i$ to
$a$, at a corrupted student sample $\vz'$. For a one-step student,
\begin{equation}
  \frac{\partial\,\KL(p_{\theta,q}\|p_{\mathrm{d},q})}{\partial u^i_b} = \E_{\lambda,\vz'}\Big[q\,\pi^i_b\Big(\Delta^i_b-\textstyle\sum_a\pi^i_a\Delta^i_a\Big)\Big].
  \label{eq:grad}
\end{equation}
\end{proposition}
Put simply, the student gives up probability on a value that the fake score favors more than the real score does,
relative to its average value. The inner sum runs over all values of each token and is computed in full, while the noise level, the latent and
the other tokens are sampled, which makes \cref{eq:grad} a Rao--Blackwellized (local-expectation) score-function
estimator~\cite{titsias2015local,liu2019raoblackwell} with the mean over values as its baseline. No relaxation~\cite{jang2017gumbel}, straight-through estimator or discriminator is needed. In training,
we apply the bracket as a fixed (stop-gradient) weight on the logits (Alg.~\ref{alg:regrn}). With more steps, the
earlier steps run without gradient, and we differentiate only the last step given its input. Matching every step hurts
(Sec.~\ref{sec:supp_diag}), because the first step is only a draft.

For bits, the gradient is a single scalar per token. When the fake score explains the observed bit better
than the real score, the student over-produces that bit, and the update moves probability to the other value
(Corollary~\ref{cor:bits}). Among $f$-divergences, only the reverse KL has a gradient that two posterior networks
can compute (Remark~\ref{rem:unique}). What is left to choose is how to weight the noise levels.

\paragraph{The window.}
\label{sec:window}
The window keeps matching away from the nearly clean end, where blank images look likely to the real score. Without
it (matching up to keep probability 0.92), 2-step ImageNet matching collapses. Training starts out healthy, and
then the samples turn into blank images (\cref{sec:components}). Near the clean end, the cross-entropy part of the
reverse KL pulls toward blank images, which the real score finds very likely (Sec.~\ref{sec:supp_window}). The fake score
that should push back is least reliable there, and the gradient is scaled down by the keep probability. Once
we cut this end, the collapse disappears for every upper end of the window from 0.3 to 0.6.

\paragraph{Random steps.}
\label{sec:randk}
A distilled student should still behave like a GRN, whose appeal is that more steps buy more quality. A $K$-step
student, however, is a specialist. It fails with fewer steps and drifts away from the data with many more
(Fig.~\ref{fig:abl_rs}a). So we draw the number of student steps anew at every iteration from a set $\Kset$
and run the whole matching iteration, fake-score updates included, with $K$-step rollouts. The draws are stratified, so
every block of $|\Kset|$ iterations visits each step count once. A single fake score serves all step counts. Trained
on samples of every count, it estimates the posterior of the mixture of their laws, and the update follows the
last-step gradient of the matching objective with this mixture in place of the student's law. A student that samples the
data law at every step count in $\Kset$ is a global minimizer (Prop.~\ref{prop:shared}).
Such a student cannot rely on a single schedule. In every iteration, its last step receives a state reached by a
different number of earlier steps, so it learns to finish whatever state it is given. A shared fake score is cheaper
than one fake score per step count, each of which would see only a fraction of the fake updates, and it is also more stable
(Fig.~\ref{fig:abl_rs}). We draw $K$ from 2--4, and from 3--6 for more refinement headroom.

\subsection{Oversampling and non-uniform sampling}
\label{sec:sampling}

\paragraph{Not all steps are equal.}
The steps of a few-step GRN carry different amounts of information and different repair loads. The first
step sees only noise. It drafts nearly the same image for every noise of a class, and across noises its predicted token
distributions differ far less than those of the last step (0.08--0.17 \vs 0.40--0.71; Tab.~\ref{tab:eh_probe}).
So the last step does most of the repair. It keeps most of the draft it is given (67--87\% of the committed
tokens, against 21--29\% by chance), so the draft fixes the layout and the last step fills in what is specific to
the sample. On the uniform grid it decides 60\% of the tokens, and it is the only step that receives gradient. The trained student also holds refinement that its own steps leave unused. Sampled with
one extra step, the 2-step GRN-B student goes from FID-10k 6.65 to 6.04, below its teacher's 6.24
(\cref{fig:components}b). Oversampling turns this unused refinement into a better judge, and non-uniform
sampling moves the repair to the step that is trained.

\paragraph{Oversampling.}
\label{sec:sbr}
Matching is only as good as its judge. Midway through training, a 2-step student has nearly matched
it. The fake and real posteriors agree closely at every noise level (0.0013--0.0032 nats per token; Tab.~\ref{tab:selffake}), and the judge
becomes the ceiling. Oversampling lifts this ceiling with the student's own extra steps.
Starting from a student and its real score, in one round we (i)~oversample the $K$-step student with $N>K$ steps,
more than it was trained for, (ii)~briefly fine-tune the current real score on these samples with GRN's own CE
objective (Eq.~(\ref{eq:sbr})), and (iii)~continue matching against the new real score, with the fake score
re-initialized from it.

This way, the student learns to do in $K$ steps what used to take $N$ (Sec.~\ref{sec:supp_fixed} gives the fixed
point). We run one round (Sec.~\ref{sec:supp_sb_impl}) with $N$ set to the student's best number of steps, which is
$N=3$ for $K=2$.

\paragraph{Non-uniform sampling.}
\label{sec:eh}
Only the last step receives gradient (\cref{sec:train}). So non-uniform sampling
gives the trained step a larger share of the path. We move a 2-step student, the base student, from the
grid $\{0,0.5\}$ to $\{0,0.25\}$ and briefly continue training and sampling on it. At equal NFE, the trained
step then covers 75\% of the path, up from half (\cref{fig:framework}). The grid has to be learned. Sampled on
$\{0,0.25\}$ without training, the base student gets worse, and $t_1=0.33$ works as well as $0.25$ on
both models (\cref{fig:components}c; Tabs.~\ref{tab:eh_full}b and~\ref{tab:eh_b}a). Trained this way, the last step adds what makes each sample
individual, and the 2-step student draws the composition of the 4-step student for the same noise (Fig.~\ref{fig:qual_eh}). A smaller
$t_1$ lets the last step decide more of the image and buys recall at the cost of precision, monotonically over six
trained values (Tab.~\ref{tab:eh_b}b).

\subsection{Refinement confidence keeping}
\label{sec:conf}
GRN's sampler keeps each predicted token with probability equal to the next grid time, so even a token that the network is sure of is
often thrown back to noise and must be found again. Refinement confidence keeping also holds on to every token whose
top predicted probability clears a threshold. It needs no extra network call, and the next step still re-predicts all
tokens. With GRN's sampling setting, 16 steps of the GRN-B teacher with keeping beat 25 steps without it, at two
thirds of the NFE (Tab.~\ref{tab:conf}). Used in every rollout of the non-uniform-sampling fine-tune and at sampling time, keeping
helps the GRN-L student as well (\cref{tab:c2i}).

\paragraph{Why it works.}
Keeping adds to GRN's random choice and leaves it intact, so each input still holds a uniformly random set of clean
tokens, as in training. Ranking tokens by confidence instead, as GRN tried~\cite{han2026grn}, gathers them in easy
regions, an input the network never sees in training, and breaks sampling. The threshold also sets when tokens are
kept, almost none early and most late. This
timing, which the threshold finds by itself, carries most of the gain on the teacher, and random keeping in the same
share per step does about as well (Tab.~\ref{tab:conf}a). With two steps, confidence also picks better tokens than chance, and training with keeping lifts the recall of
\method-L (Tab.~\ref{tab:c2i_full}).

\section{Experiments}
\label{sec:experiments}

\begin{figure*}[t]
  \centering
  \IncludeOrPH[width=\linewidth]{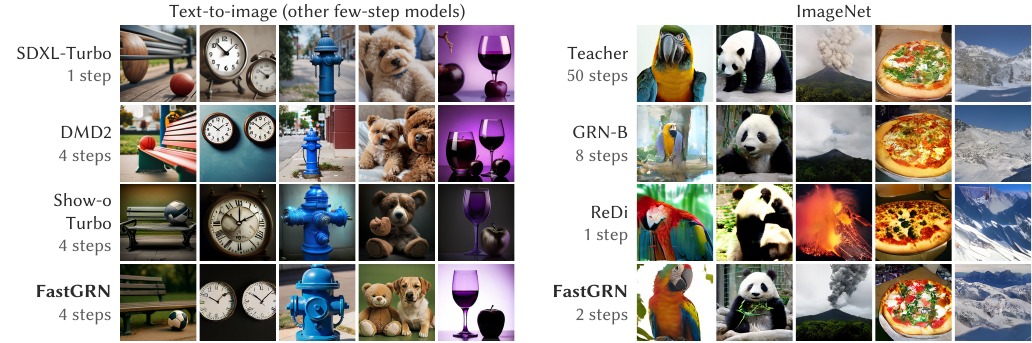}{\linewidth}{6.9cm}{\textbf{Visual comparison placeholder}}
  \vspace{-1.4em}
  \caption{\textbf{Few-step generators side by side.} Left: few-step text-to-image models as
  released~\cite{sauer2024add,yin2024dmd2,xu2025showoturbo} and \method on GenEval prompts (\emph{a bench and a sports
  ball}, \emph{two clocks}, \emph{a blue fire hydrant}, \emph{a dog right of a teddy bear}, \emph{a purple wine glass and
  a black apple}), all shown at $256^2$. Right: ReDi~\cite{redi2025} as released, and the GRN-B rows with one frozen noise per column.}
  \label{fig:compare}
\end{figure*}

\begin{figure*}[t]
  \centering
  \IncludeOrPH[width=\linewidth]{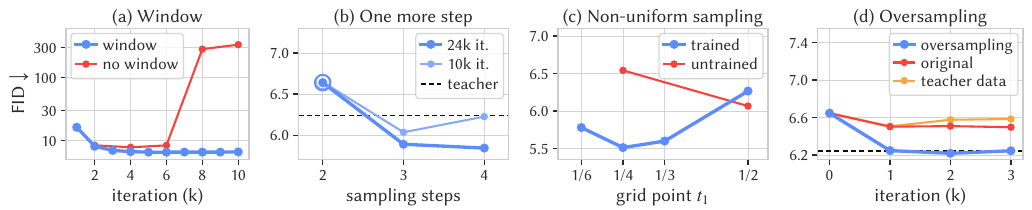}{\linewidth}{3.8cm}{\textbf{Component figure placeholder}}
  \vspace{-1.6em}
  \caption{\textbf{Each component pays off} (ImageNet $256^2$, FID-10k, GRN-B except (c); black dashed: teacher).
  \textbf{(a)}~Without the window, 2-step training collapses. \textbf{(b)}~One step more than it was trained for
  brings the 2-step student below its teacher (ring: trained steps). \textbf{(c)}~The non-uniform grid of GRN-L helps
  only when the student is trained on it. \textbf{(d)}~Of three judges, only oversampling brings the 2-step student to
  the teacher.}
  \label{fig:components}
\end{figure*}

\subsection{Setup}
\label{sec:setup}
\paragraph{ImageNet.}
We distill the released GRN-B (130M) and GRN-L (458M) on ImageNet~\cite{deng2009imagenet} $256^2$ from teacher
samples only. The teachers use the setting of the GRN paper, with 50 steps and CFG on $t\in(0.4,1)$ for 79 NFE.
Re-evaluated with our pipeline, they score an FID-50k of 3.73 (GRN-B) and 2.65 (GRN-L), compared with 3.56 and 2.64
in the GRN paper, and all comparisons use our numbers. We fine-tune GRN on teacher samples with GRN's
corruption, which bakes the guidance into the weights. This fine-tuned GRN serves as the real score and initializes
the fake score. The paired-noise GRN initializes the student. Windowed matching uses $\qmax=0.4$ and $\tau=0.8$ on both models, and
Sec.~\ref{sec:supp_impl} gives all recipes. We report FID-50k~\cite{heusel2017fid} and
IS~\cite{salimans2016improved} with the teacher's protocol, together with precision and
recall~\cite{kynkaanniemi2019prec} against 50k teacher samples. Studies use FID-10k.

\paragraph{Text-to-image.}
We use the public GRN-T2I 2B checkpoint, which works on bits at $256^2$ and samples with 50 steps and CFG at every
step for 100 NFE. With one teacher sample per prompt, we train LoRA~\cite{hu2022lora} adapters on the frozen base for
the fine-tuned GRN, the paired-noise GRN, the student and its fake score (Sec.~\ref{sec:supp_impl}). We report
GenEval~\cite{ghosh2023geneval}, FID-T, HPSv2.1~\cite{wu2023hpsv2} and LPIPS~\cite{zhang2018lpips} diversity, where
FID-T is the FID to teacher images for the same COCO~\cite{lin2014coco} captions.

\paragraph{Component studies.}
We use GRN-B, and GRN-L too for non-uniform sampling and refinement confidence keeping. We compare the
oversampling judge with the original judge, the fine-tuned GRN, and with a \emph{teacher-data judge}, which is the
original judge fine-tuned like the oversampling judge on the teacher samples it was trained on.

\InputOrPH{tables/tab_c2i_main.tex}{\TablePH{tab:c2i}{Class-conditional ImageNet $256^2$}{3.2cm}}
\InputOrPH{tables/tab_compare.tex}{\TablePH{tab:compare}{Few-step generators on ImageNet $256^2$}{4.0cm}}

\subsection{Main results}
\label{sec:c2i}
\label{sec:speed}
\paragraph{ImageNet.}
We give the main results in \cref{tab:c2i}. With GRN-B, two refinement steps match the teacher's FID-50k of 3.73 with
40$\times$ fewer NFE. Every component lowers FID, and none adds a network call at sampling time. With 3 steps, the same
student goes below the teacher. GRN-L shows the same trend. Non-uniform sampling and refinement confidence keeping close
the gap between the 2-step and the 4-step student, oversampling lowers FID further, and with 3 steps the student goes
below its teacher. The 3-step students use 3 NFE, 26 times fewer than their teachers. Including decoding, an
image takes tens of milliseconds on an RTX 4090, compared with about a second for the teacher
(Tab.~\ref{tab:speed}).
The other few-step routes from the same teacher fall far behind (\cref{tab:c2i}). Even with a tuned temperature, the
teacher's own sampler needs 12 steps, 19 NFE, to reach this level (Tab.~\ref{tab:c2i_full}).

\paragraph{Other few-step students.}
In \cref{tab:compare} we compare \method with other few-step students of discrete generators. Without a GAN loss, only
\method-B matches its teacher's FID with two network calls, and Di4C needs four guided steps. Soft-Di[M]O goes below
its teachers with a GAN loss on real images. \method also has the lowest FID among the students of MaskGIT and VAR
teachers. Tab.~\ref{tab:compare_ctx} adds more students, native samplers and continuous generators.

\begin{figure}[t]
  \centering
  \IncludeOrPH[width=\linewidth]{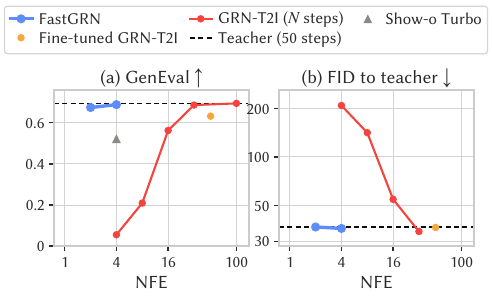}{\linewidth}{5.9cm}{\textbf{Text-to-image plot placeholder}}
  \vspace{-1.6em}
  \caption{\textbf{Text-to-image with few steps} (GRN-T2I 2B, $256^2$). GRN-T2I ($N$ steps): the teacher's own sampler
  with CFG at every step. Show-o Turbo as reported. Dashed: the teacher with 50 steps, in (b) against its own resamples. Notes:
  Sec.~\ref{sec:supp_tabnotes}.}
  \label{fig:t2i}
\end{figure}

\InputOrPH{tables/tab_t2i_compare.tex}{\TablePH{tab:t2i_compare}{Few-step text-to-image students}{3.6cm}}
\paragraph{Text-to-image.}
On text-to-image, the teacher's own sampler falls apart with few steps. Its GenEval drops at 8 steps and collapses at
4 (\cref{fig:t2i}). With the window alone, the 4-step \method student nearly keeps the teacher's GenEval (0.689 \vs
0.695), a level that the teacher's own sampler approaches only with 8$\times$ the NFE. The student is as close to the
teacher as the teacher's own resamples in FID-T and keeps most of its diversity (Tab.~\ref{tab:t2i_full}).
In \cref{tab:t2i_compare} we compare it with the few-step students of other discrete generators. With two steps, \method
keeps 97\% of the teacher's GenEval and already scores above all of them, including one that adds GAN and reward losses and test-time search. Only the
reward-tuned students keep more of their teacher's score.

\paragraph{Qualitative comparison.}
In \cref{fig:compare} (left) we put \method next to the released few-step students of SDXL and Show-o. They part
ways when a prompt binds attributes or relations. SDXL-Turbo and DMD2
paint the black apple purple, Show-o Turbo draws one clock where the prompt asks for two, and SDXL-Turbo and Show-o
Turbo each draw a single animal for a dog right of a teddy bear. \method follows every prompt. On ImageNet (right),
the teacher's own sampler with 8 steps loses fine structure and turns textures blotchy, and
the 1-step ReDi~\cite{redi2025} oversaturates colors and smears shapes. \method restores sharp texture with 2 steps (Fig.~\ref{fig:compare_t2i_a}). Oversampling keeps each sample's composition
and sharpens local detail (Fig.~\ref{fig:qual_sb}), while non-uniform sampling
changes the composition itself, so the 2-step student draws the layout of the 4-step one (Fig.~\ref{fig:qual_eh}).

\subsection{Component study}
\label{sec:components}
\paragraph{Window and random steps (\cref{sec:train}).}
Without the window, the 2-step student looks healthy early on, then its samples turn blank and FID explodes. With the window, training stays stable (\cref{fig:components}a). The window also helps where nothing
collapses, such as on the 4-step student (Tab.~\ref{tab:abl_c2i}).
A fixed-step student is a specialist and drifts with more steps. With steps drawn at random from 2--4, the curve
becomes flat from 4 to 8 steps, and at 4 steps this student beats the fixed 4-step student in half the iterations.
With 3--6, it keeps improving and reaches the teacher's FID at 8 (Fig.~\ref{fig:abl_rs}).

\paragraph{One more step.}
Only the last step is trained, so a 2-step student still has refinement left. A third sampling step brings it below
its teacher's FID, and a fourth adds little (\cref{fig:components}b). Oversampling turns this unused step into a
better judge, and the 3-step rows of \cref{tab:c2i} use the extra step to beat both teachers (GRN-L after 2k more
iterations with a decaying learning rate).

\paragraph{Oversampling and non-uniform sampling (\cref{sec:sampling}).}
The oversampling judge works because it has seen the student's own refined samples. One round brings the 2-step
student to the teacher's level with 2 NFE, compared with 79 for the teacher (\cref{fig:components}d). The original judge and the teacher-data judge stop well short.
On GRN-L, a round on the 2-step student with keeping, whose judge is fine-tuned on the student's own 3-step samples,
lowers FID further (\cref{tab:c2i}).

On GRN-B, a short fine-tune on $\{0,0.25\}$ improves the
\tagW+\tagSB{} student, while the same number of iterations on $\{0,0.5\}$ does not, and on the standard grid
this student even drops below the teacher's FID (Tab.~\ref{tab:eh_b}a). GRN-L behaves the same way
(Tab.~\ref{tab:eh_full}). Sampled on the non-uniform grid without the training, the student gets worse, so what
matters is training on that grid, and $t_1$ near $0.25$ works best (\cref{fig:components}c).

\paragraph{Refinement confidence keeping (\cref{sec:conf}).}
Keeping pays off once the student learns with it. Without training, the students' lower threshold hurts. Trained with keeping in every rollout, the 2-step \method-L reaches the FID of our 4-step student at 2 NFE,
mostly through recall, while \method-B stays at its teacher's FID (Tab.~\ref{tab:conf}).

\section{Conclusion}
\label{sec:conclusion}
Few-step GRNs fail because of a joint error between tokens, and distribution matching on the student's own samples
reduces this error. For uniform-state tokens, the gradient of distribution matching needs no sampled-token gradient
estimator, because it is computed directly from two posterior networks. \method builds few-step GRNs on this
gradient in three parts. A window keeps matching stable, and random steps train one student for 4 to 8 steps.
Oversampling and non-uniform sampling turn the student's own extra steps into a better judge and lengthen the trained
step. Refinement confidence keeping holds on to the tokens the student is already sure of. A 2-step GRN-B student
matches the FID of its 79-NFE teacher, and with one more sampling step the students of both GRN-B and GRN-L go
below their teachers. The same parts apply to any generator trained with uniform-state corruption, and bit-level text-to-image and
video models are the natural next step.

{
    \small
    \bibliographystyle{ieeenat_fullname}
    \bibliography{references}
}

\clearpage
\setcounter{section}{0}
\setcounter{figure}{0}
\setcounter{table}{0}
\setcounter{equation}{0}
\setcounter{lemma}{0}
\setcounter{proposition}{0}
\setcounter{corollary}{0}
\setcounter{remark}{0}
\setcounter{definition}{0}
\setcounter{algorithm}{0}
\renewcommand{\thesection}{S\arabic{section}}
\renewcommand{\thetable}{S\arabic{table}}
\renewcommand{\thefigure}{S\arabic{figure}}
\renewcommand{\theequation}{S\arabic{equation}}
\renewcommand{\thelemma}{S\arabic{lemma}}
\renewcommand{\thecorollary}{S\arabic{corollary}}
\renewcommand{\theremark}{S\arabic{remark}}
\renewcommand{\thealgorithm}{S\arabic{algorithm}}
\makeatletter
\def\theHsection{S\arabic{section}}
\def\theHsubsection{\theHsection.\arabic{subsection}}
\def\theHsubsubsection{\theHsubsection.\arabic{subsubsection}}
\def\theHparagraph{\theHsubsubsection.\arabic{paragraph}}
\def\theHtable{S\arabic{table}}
\def\theHfigure{S\arabic{figure}}
\def\theHequation{S\arabic{equation}}
\def\theHlemma{S\arabic{lemma}}
\def\theHproposition{S\arabic{proposition}}
\def\theHcorollary{S\arabic{corollary}}
\def\theHremark{S\arabic{remark}}
\def\theHdefinition{S\arabic{definition}}
\def\theHalgorithm{S\arabic{algorithm}}
\def\theHALG@line{\thealgorithm.\arabic{ALG@line}}
\makeatother
\renewcommand{\dbltopfraction}{0.85}
\raggedbottom
\maketitlesupplementary

\section{Proofs}
\label{sec:supp_proofs}
\renewcommand{\theproposition}{S\arabic{proposition}}

Notation follows the main paper. $k_q$ is the uniform-state kernel of Eq.~(\ref{eq:kernel}) at keep probability
$q$, and $\nu=(1-q)/\Vsz$. The concrete score $s(\vz,i,a)$ of a law is its probability at $\vz$ with token $i$ set to
$a$, divided by its probability at $\vz$. The $K$-step student $\Gs$ has law $p_\theta$, and $p_{\theta,q}$ is that law
corrupted by $k_q$. $\Rs$ and $\Fs$ are the real and fake scores with posterior marginals $D_R$ and $D_F$.

\subsection{Why few steps fail}
\label{sec:supp_tc}

\begin{lemma}[Redundancy of per-token interval maps]
\label{lem:redundancy}
If the corruptions at two noise levels are coupled token by token given the clean image, as in uniform-state
diffusion, the CE-optimal per-token interval predictor, which maps a token's state at a noisier time to its state at
a cleaner one as MeanFlow~\cite{geng2025meanflow} does for continuous data, is a fixed function of the posterior
marginals that GRN already outputs. Sampling with it is ancestral sampling with GRN's own posterior.
\end{lemma}
\begin{proof}
Given the clean image, the two states of a token depend only on its clean value, so the next state is the posterior
marginal passed through a known per-token bridge, which a CE-trained predictor recovers.
\end{proof}

\begin{proposition}[Few-step error is a joint error]
\label{prop:tc}
Among factorized predictors of the image given a state $\vz$, the expected CE is minimized by the product of the
posterior marginals, and the remaining error is
\begin{equation}
  \E_{\vz}\,\KL\big(p(\cdot\mid\vz)\,\big\|\,{\textstyle\prod_i} p(x^i\mid\vz)\big) \;=\; \E_{\vz}\,\TC(\vx\mid\vz),
  \label{eq:tc}
\end{equation}
the conditional total correlation~\cite{watanabe1960tc}.
\end{proposition}
\begin{proof}
The CE of a factorized predictor splits over tokens, and each term is smallest at the true marginal (Gibbs'
inequality). The KL from a joint law to the product of its marginals is the sum of the marginal entropies minus the
joint entropy. From pure noise the sampler outputs the product of the data marginals, at distance $\TC(\vx)$ from
the data.
\end{proof}

\begin{corollary}[Informative noise]
\label{cor:paired}
If training pairs of noise and image are coupled so that the images follow the data law, the KL from the data to the
CE-optimal one-step sampler is at most the average conditional total correlation given the noise. It is zero when the
noise determines the image, and $\TC(\vx)$ for noise independent of the image.
\end{corollary}
\begin{proof}
Both laws are averages over the noise, of the conditional law and of the product of its marginals. Joint convexity
of the KL and \cref{prop:tc} give the bound.
\end{proof}
\noindent Paired-noise fine-tuning realizes this coupling. The student's last step is factorized only given the noise
and the earlier states, so its law is a mixture of product laws that distribution matching can shape.

\subsection{The matching gradient}
Corruption acts token by token, so the concrete score of the corrupted law averages the kernel ratio
$k_q(a\mid b)/k_q(z^i\mid b)$ over the posterior of the clean value $b$. With $D^i$ the posterior marginal of token
$i$ and $a\ne z^i$,
\begin{equation}
  s(\vz,i,a)=1+\frac{q\,D^i(a)}{\nu}-\frac{q\,D^i(z^i)}{q+\nu}.
  \label{eq:delta}
\end{equation}
The windowed matching objective of Sec.~\ref{sec:wdm} is
\begin{equation}
  \mathcal{L}(\theta) \;=\; \E_{q\sim\rho}\big[\KL\big(p_{\theta,q}\,\big\|\,p_{\mathrm{d},q}\big)\big].
  \label{eq:wdm}
\end{equation}
\begin{proof}[Proof of Proposition~\ref{prop:grad}]
The total mass is fixed, so the change of the KL is the change of the corrupted student law weighted by the log
density ratio. For $K=1$, token $i$ of the corrupted student takes value $a$ with probability $q\pi^i(a)+\nu$, so a
change of $\pi^i$ only moves mass between states that differ in token $i$. The log density ratio thus enters through
its change when token $i$ is set to $a$, which is $\Delta^i_a$, the difference of the log concrete scores
(\cref{eq:delta}) computed with $D_F$ and $D_R$. The softmax subtracts the student's own average of $\Delta^i$. For
$K>1$ we ignore how the input of the last step depends on $\theta$ (a truncated gradient).
\end{proof}
\noindent Training applies it through the surrogate $\sum_{i,b}\sg(g^i_b)\,l^i_b/\tau$, where $g$ is the bracket of
Eq.~(\ref{eq:grad}) (\cref{alg:regrn}).

\begin{corollary}[Binary tokens]
\label{cor:bits}
For bits ($\Vsz=2$), $\Delta^i$ vanishes at the observed bit $z$, and with $c=\big(\tfrac{1+q}{2}\big)^2$ the
gradient with respect to the logit difference $d^i=u^i_1-u^i_0$ is
\begin{equation}
  \frac{\partial\,\KL}{\partial d^i}=\E\Big[q\,\pi^i_0\pi^i_1\,(1-2z)\,\log\frac{c-q\,D_F^i(z)}{c-q\,D_R^i(z)}\Big].
  \label{eq:bits}
\end{equation}
\end{corollary}
\begin{proof}
With $q+2\nu=1$, \cref{eq:delta} gives the score of flipping the bit as $(c-qD^i(z))/(\nu(q+\nu))$, and the
two-logit softmax turns $\Delta$ into the gradient on $d^i$.
\end{proof}
\noindent If the fake score explains the observed bit better than the real score, the student over-produces it and
the update moves probability to the other value.

\begin{remark}[Why reverse KL]
\label{rem:unique}
For an $f$-divergence the bracket of Eq.~(\ref{eq:grad}) becomes $f'(cr)-f'(r)$, with $r$ the density ratio and
$c$ the ratio of concrete scores that two posterior networks provide. It depends on $c$ alone for every $r$ iff
$f'(e^x)$ is affine (Cauchy's functional equation), \ie only for the reverse KL up to scale. Other divergences need
absolute density ratios, \eg from a discriminator, so the free design choice is the weighting $\rho$, which the
window changes.
\end{remark}

\subsection{Random steps and oversampling}
\label{sec:supp_proof_shared}
\label{sec:supp_fixed}

\begin{proposition}[Shared fake score]
\label{prop:shared}
With random steps (Sec.~\ref{sec:randk}) one fake score serves all $K\in\Kset$, so $D_F$ is the posterior of the
corrupted mixture $\bar p_\theta$ of the $K$-step laws. Eq.~(\ref{eq:grad}) applied to $K$-step samples and averaged
over $K$ uniform on $\Kset$ is the last-step gradient of \cref{eq:wdm} with $\bar p_\theta$ in place of $p_\theta$. A
student whose every $K$-step law is the data law is a global minimizer.
\end{proposition}
\begin{proof}
The change of the mixture KL is the average over $K$ of the changes of the $K$-step laws, weighted by the log ratio of
the mixture to the data, and for each $K$ the proof above applies with the concrete score of the mixture. The KL is
zero iff the mixture equals the data law (the kernel is invertible for $q>0$), which holds if every $K$-step law does.
\end{proof}
\noindent Stratified draws make $K$ exactly uniform within every block of $|\Kset|$ iterations
(\cref{sec:supp_rs_impl}), so the CE optimum of the shared fake score is this posterior.

An oversampling round (Sec.~\ref{sec:sbr}) samples the $K$-step student $\theta_r$ with $N>K$ steps, briefly
fine-tunes the real score on these samples with GRN's objective,
\begin{equation}
  \Rs_{r+1}=\mathrm{FT}_{\Rs_r}\Big[\min_{\Rs}\ \E_{\vx\sim p^{N}_{\theta_r}}\,\mathrm{CE}(\Rs;\vx)\Big],
  \label{eq:sbr}
\end{equation}
with $\mathrm{FT}_{\Rs_r}$ a short fine-tune from $\Rs_r$, and continues windowed matching against $\Rs_{r+1}$ with
the fake score re-initialized from it.

\begin{remark}[Fixed point]
\label{rem:supp_fixed}
With exact minimization, a round leaves $(\theta,\Rs)$ unchanged iff $p^{K}_{\theta}=p_{\Rs}=p^{N}_{\theta}$, where
$p_{\Rs}$ is the law whose posteriors $\Rs$ gives. The judge learns the $N$-step law, and since the kernel is
invertible, matching then makes the $K$-step law equal to it. As in expert iteration~\cite{anthony2017exit}, the slow
sampler is the expert and the fast one the apprentice, here taught through the judge.
\end{remark}

\paragraph{The window keeps the minimizer.}
For $q>0$ the per-token kernel is invertible, so each KL term of \cref{eq:wdm} vanishes only at the data law, and so
does any average over $q\in[\qmin,\qmax]$ with $\qmin>0$.

\section{Implementation details}
\label{sec:supp_impl}

\paragraph{Teachers and data.}
Times are wall-clock on A100, L40S and RTX 4090 GPUs. The teachers are the released GRN-B (130.9M parameters,
16-level tokens on a $16{\times}16{\times}16$ map) and GRN-L (458.5M), sampled as in the GRN paper with 50 steps,
CFG on $t\in(0.4,1)$, scale 2.4 / 2.0 and temperature 1.33 / 1.30 for B / L (79 NFE), and uniform class labels. For
GRN-B we draw 200k teacher samples for the fine-tuned GRN and 600k pairs for the paired-noise GRN (2.9\,h each). For
GRN-L, 200k samples serve both.

\paragraph{Fine-tuned and paired-noise GRN.}
Both start from the released EMA weights and train with GRN's CE on all tokens, without label dropout, with
$t\sim0.6\,\mathrm{sigmoid}(\mathcal{N}(-0.8,0.8))+0.4\,\mathcal{U}(0,0.995)$, AdamW~\cite{loshchilov2019adamw}
($\beta=(0.9,0.95)$, no weight decay), learning rate $5\cdot10^{-5}$ after 1k warm-up iterations, gradient clipping
1.0, bf16 autocast and batch 128. The fine-tuned GRN keeps GRN's corruption with fresh noise and trains for 15k
iterations. The paired-noise GRN fills dropped tokens with the sample's own initial noise. On GRN-B it starts from the
fine-tuned GRN and trains for 40k iterations, and on GRN-L for 15k.

\paragraph{Windowed matching on ImageNet.}
The student starts from the paired-noise GRN (EMA weights) and the fake score from the fine-tuned GRN, which is also
the frozen real score. Student and fake updates ($n_{\text{fake}}=5$) use batch 64 and learning rates $10^{-5}$ /
$2\cdot10^{-5}$. We draw $q\sim0.5\,\mathrm{sigmoid}(\mathcal{N}(-0.8,0.8))+0.5\,\mathcal{U}(0,1)$ clamped to
$[0.02,\qmax]$ with $\qmax=0.4$ (0.92 without the window), which puts 46\% of the draws at $\qmax$. The fake score
trains at all levels with the $t$-law above. The student uses $\tau=0.8$ and EMA 0.999, trains for 10k iterations
and samples with GRN's frozen-noise sampler on $t_s=s/K$ without CFG. A student takes up to 3.8\,h (4 steps), and
the whole GRN-B and GRN-L pipelines take about 15 and 19 GPU-hours.

\paragraph{Random steps.}
\label{sec:supp_rs_impl}
Iterations are grouped in blocks of $n=|\Kset|$. Each block is a permutation of $\Kset$ from a counter-based random
generator keyed by the block index, and iteration $j$ takes entry $j \bmod n$. The whole iteration, its five fake
updates included, uses $K_j$-step rollouts (\cref{alg:regrn}), and one fake score takes all updates. The random-step
students and the fixed-step students they are compared with train for 6k iterations and are evaluated with 1--6 and 8
steps. The per-$K$ variant keeps one fake score for each $K$ and updates it only when $K$ is drawn ($5/n$ updates per
student update). The variant with 15 fake updates restores five per fake score at 2.6 times the cost.

\paragraph{Oversampling.}
\label{sec:supp_sb_impl}
A round (i)~draws 200k samples, and 5k more for validation, from the student's EMA weights with $N$ steps and
$\tau=0.8$, (ii)~fine-tunes the judge, the fine-tuned GRN (EMA 0.9995), on them for 4k iterations with GRN's CE,
batch 128 and learning rate $2\cdot10^{-5}$ after 200 warm-up iterations, and takes its EMA 0.999 weights as the new
real score, and (iii)~continues windowed matching for 3k iterations against it with the original recipe, the fake
score re-initialized from the judge. The 2-step student uses $N=3$, its best step count. The original-judge
continuation runs step (iii) alone against the unchanged judge, and the teacher-data judge runs steps (ii) and (iii)
on the 200k teacher samples the judge was trained on. The same round is applied to the 6k $K{=}3$--$6$ student
($N=8$), the fixed 4-step student ($N=5$) and the GRN-L $K{=}3$--$6$ student ($N=5$). For the 2-step GRN-B student a
round takes 1.39 GPU-hours on one RTX 4090, 42\% of the student's own training, and sampling and the judge fine-tune
add only 12\%.

\paragraph{Non-uniform sampling.}
\label{sec:supp_eh_impl}
Each run continues a 2-step student from its live weights for 2k iterations with the recipe of its model and the
gradient on the last step only, and trains and samples on the same grid $\{0,t_1\}$, with $t_1=0.25$ for non-uniform
sampling and $0.5$ for the standard grid. With $t_1=0.25$ the trained last step covers 75\% of the path. On GRN-L the
base is the 10k \tagW{} student, and the fine-tuned GRN-L (EMA 0.9995) is the real score and the fake initialization
(2.2\,h per run). On GRN-B the base is the \tagW+\tagSB{} student, with its oversampling judge as the real score and a
fresh fake score (1.5\,h). The \tagW{} runs of \cref{tab:eh_b}b continue the 10k \tagW{} student against the
fine-tuned GRN.

\begin{algorithm}[t]
\caption{One \method training iteration (windowed matching with random steps)}
\label{alg:regrn}
\small
\begin{algorithmic}[1]
\Require student $\Gs$ (init.\ paired-noise GRN), fake score $\Fs$ (init.\ fine-tuned GRN), frozen real score $\Rs$, step set $\Kset$ (fixed $K$: $K$ only), window $[\qmin,\qmax]$, iteration index $j$
\State $K\gets\sigma_b(j \bmod |\Kset|)$, $\sigma_b$ a random permutation of $\Kset$, $b=\lfloor j/|\Kset|\rfloor$\Comment{stratified}
\For{$m=1,\dots,n_{\text{fake}}$}\Comment{fake-score updates}
  \State $\vx\gets$ $K$-step rollout of $\Gs$ from fresh $(\veps,\vc)$, no grad
  \State $t\sim$ GRN's $t$-law, $\vz\sim k_t(\cdot\mid\vx)$
  \State step on $\mathrm{CE}(\Fs(\vz,t,\vc),\vx)$
\EndFor
\State rollout from $(\veps,\vc)$: steps $1..K{-}1$ no grad; last-step logits $l$, $\hat\vx\sim\softmax(l/\tau)$
\State $q\sim\rho$ clamped to $[\qmin,\qmax]$; \ $\vz'\sim\kq(\cdot\mid\hat\vx)$
\State $D_F\gets\Fs(\vz',q,\vc)$, \ $D_R\gets\Rs(\vz',q,\vc)$ \Comment{no grad}
\State $g\gets q\,\pi\,(\Delta-\textstyle\sum_a\pi_a\Delta_a)$ via Eqs.~(\ref{eq:delta}) and (\ref{eq:grad})
\State step on $\sum\sg(g)\,l/\tau$; update EMA of $\theta$
\end{algorithmic}
\end{algorithm}

\paragraph{Text-to-image.}
The teacher is the public GRN-T2I 2B checkpoint at $256^2$ ($16{\times}16$ positions $\times$ 256 bits $=65{,}536$
binary tokens), frozen in bf16, with 50 steps, CFG 4.5 and $\tau=1.1$ (100 NFE), and its HBQ decoder. The pseudo
data are one teacher sample for each of 50k prompts from DenseFusion, COCO-2014 train and DiffusionDB, without the
evaluation prompts. Every trained model is a rank-64 LoRA on the 7 linear layers of all 28 blocks (89.5M parameters)
of one frozen base. The fine-tuned and the paired-noise GRN train for 5k iterations at batch 64. Matching follows the
ImageNet setup (the guided variant uses the CFG-combined base as real score after 200 fake-only updates) with batch
32, learning rates $5\cdot10^{-5}$ / $10^{-4}$ after 200 warm-up iterations, $\tau=1$ and EMA 0.999. One iteration
takes 14 / 17.5 / 25\,s for 1 / 2 / 4 steps. All text-to-image students use windowed matching without the other
parts (\tagW).

\paragraph{Evaluation.}
ImageNet FID and IS follow the teacher's protocol (bf16 generation, fp32 HBQ decoding) and reference statistics.
Precision and recall use $k=3$ Inception neighbours against 50k teacher samples drawn with separate noise. FID-10k,
used in the studies, runs 2.3--2.9 above FID-50k. FID-T is the FID of a student's samples to the 50k teacher samples.
For text-to-image we report GenEval~\cite{ghosh2023geneval} (553 prompts, four images each, Mask2Former
detector~\cite{cheng2022mask2former}) with paired differences to the teacher, which scores 0.695 at $256^2$ (GRN's
0.76 is measured at $1024^2$ with a prompt rewriter). Here FID-T is the FID to teacher images for 1k
COCO~\cite{lin2014coco} captions, where a re-sampled teacher scores 36.6. We also report
CLIP-L/14~\cite{radford2021clip}, HPSv2.1~\cite{wu2023hpsv2} and same-prompt LPIPS~\cite{zhang2018lpips} diversity
relative to the teacher. Batch-1 latency is the median of 20--30 runs after warm-up with \texttt{torch.compile} on one
RTX 4090 and includes the HBQ decoder (34\,ms).

\paragraph{Notes to the main-paper tables and Fig.~\ref{fig:t2i}.}
\label{sec:supp_tabnotes}
\emph{Table~\ref{tab:c2i}.} GRN-B, 8 steps: the teacher's own sampler with fewer steps, at the temperature with the lowest FID-10k for 8 steps, 0.7 (GRN uses 1.33 with 50 steps; CFG on $t\in(0.4,1)$ adds network calls); Fine-tuned and paired-noise GRN: GRN fine-tuned on teacher samples with independent and with paired noise; matching without window: distribution matching without the window, at its best checkpoint (4k iterations) before it collapses. Bold: best student FID. ms: batch-1 latency with decoding; the paired-noise GRN and matching without window share it with the 2-step students (same network), the other baselines are not timed. Tags: \tagW{} paired-noise initialization and windowed distribution matching (10k iterations); \tagRS{} random steps over $K{=}2$--$4$ (6k); \tagSB{} one oversampling round (+3k); \tagEH{} non-uniform sampling, +2k iterations on $\{0,0.25\}$; \tagCK{} refinement confidence keeping in every rollout of that fine-tune and when sampling. GRN-B with 3 steps: the 2-step \tagW+\tagSB+\tagEH{} student sampled with 3 steps on $\{0,1/3,2/3\}$ (no training; 3-step latency not measured). GRN-L, \tagW+\tagEH+\tagCK+\tagSB: one oversampling round on the 2-step \tagW+\tagEH+\tagCK{} student, with the judge fine-tuned on its own 3-step samples (+2k); GRN-L with 3 steps: the \tagW+\tagEH+\tagCK{} student after 2k more iterations with a learning rate decaying to 5\% over the last 1k, sampled with 3 steps on $\{0,0.25,0.5\}$ without keeping (3-step latency not measured; Fig.~\ref{fig:teaser} places the 3-step points on the line through the measured 1-, 2- and 4-step latencies of the same network). All students use EMA 0.999.

\emph{Table~\ref{tab:compare}.} Numbers reported by the cited papers, each with its own tokenizer and its own teacher, so students are compared by $\times$T.\ (bold: lowest without a GAN loss; teacher FID: MaskGIT 6.60, MaskBit 1.66, VAR-d16 4.15 for DD2 and 4.19 for DD1, VAR-d20 3.40, Di4C's 8-step MaskGIT 6.57). $^g$Trained with a GAN loss on real data. $^c$Sampling steps as reported; these methods use classifier-free guidance at inference. $^\dagger$Size not stated in the paper; these students start from the PyTorch MaskGIT of Besnier and Chen~\cite{besnier2023maskgitpytorch}, whose size Di[M]O reports. SDTT: three rounds, as run by ReDi. DAG: four network evaluations, initialized from ReDi. \regrn: rows of Tab.~\ref{tab:c2i}, the 2-step \tagW+\tagSB+\tagEH{} (GRN-B) and \tagW+\tagEH+\tagCK+\tagSB{} (GRN-L) students and the GRN-L student with 3 steps. More students of MaskGIT, VAR and LlamaGen teachers (SDTT, ReDi, Di4C, DD1, DD2), native discrete samplers and few-step generators in continuous latent spaces (with parameters and IS): Tab.~\ref{tab:compare_ctx}.

\emph{Figure~\ref{fig:t2i}.} All students are LoRA adapters trained with windowed matching alone (\tagW): 4 steps with $q_{\max}{=}0.40$ for 2k iterations, and 2 steps with $q_{\max}{=}0.55$ for 2k iterations, then 1k at $q_{\max}{=}0.95$. FID-T: FID to 1k teacher images for the same captions, \ie closeness to the teacher; the dashed line in (b) is the teacher against a second teacher sample set, \ie the teacher's own noise level. GRN-T2I ($N$ steps): the teacher's own sampler with fewer steps (CFG at every step, $2N$ NFE). Fine-tuned GRN-T2I: GRN-T2I fine-tuned on teacher samples, 50 steps without CFG. All values, HPS, sub-task scores, CLIP, the paired GenEval difference to the teacher, diversity and more rows: Tab.~\ref{tab:t2i_full}. Show-o Turbo: the 4-step student of the discrete Show-o, GenEval as reported (at $256^2$, without CFG). All reported students, continuous ones included: Tab.~\ref{tab:t2i_ctx}.

\emph{Table~\ref{tab:t2i_compare}.} GenEval overall as reported by the cited papers; each student is divided by its own teacher in the same paper (MaskGen-L with 16 steps 0.48, as run by Soft-Di[M]O and FPD; Show-o with 16 steps and CFG 0.591 at $256^2$ and 0.674 at $512^2$; GRN-T2I with 50 steps and CFG 0.695). Bold: highest GenEval, and highest $\times$T.\ without a GAN or reward loss. $^g$GAN loss; $^r$reward fine-tuning; $^s$TTS: test-time search guided by reward models, with many network calls per image. Show-o Turbo and \regrn{} sample without CFG, so steps equal network calls. \regrn: the 2- and 4-step students of Fig.~\ref{fig:t2i}. Students of Meissonic, Show-o Turbo with 8 steps and continuous few-step generators: Tab.~\ref{tab:t2i_ctx}.

\section{Additional results}
\label{sec:supp_results}

\subsection{Random steps and oversampling}
\label{sec:supp_components}
\label{sec:supp_randk}
\begin{figure*}[t]
  \centering
  \IncludeOrPH[width=\linewidth]{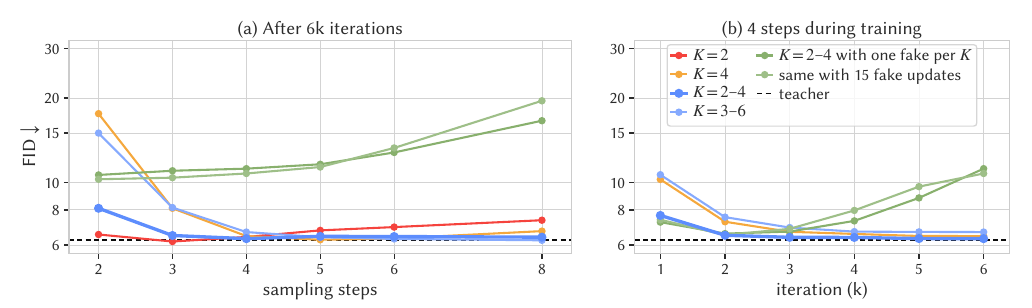}{\linewidth}{5.2cm}{\textbf{Random steps placeholder}}
  \vspace{-1.6em}
  \caption{\textbf{Random steps flatten the curve} (GRN-B, FID-10k). \textbf{(a)}~After 6k iterations, by sampling
  steps. \textbf{(b)}~4-step FID during training. One fake per $K$: a separate fake score for every step count.}
  \label{fig:abl_rs}
\end{figure*}
\InputOrPH{tables/tab_sb_full.tex}{\TablePHwide{tab:sb_full}{Oversampling: full results}{5.0cm}}
\begin{figure*}[t]
  \centering
  \IncludeOrPH[width=\linewidth]{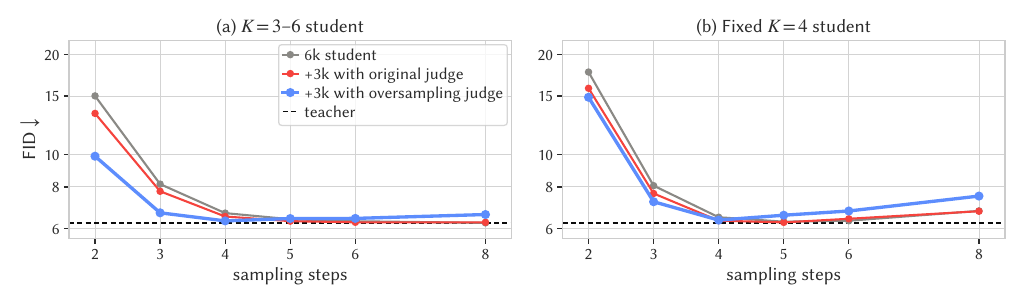}{\linewidth}{5.0cm}{\textbf{Oversampling on 4-step students placeholder}}
  \vspace{-1.6em}
  \caption{\textbf{One oversampling round on the 4-step students} (GRN-B, FID-10k; 3k iterations of matching against
  each judge).}
  \label{fig:abl_os4}
\end{figure*}

\paragraph{Random steps.}
\Cref{fig:abl_rs} compares the random-step students of Sec.~\ref{sec:randk} with fixed-step students.
Random steps trade the sharp optimum of a specialist for a flat curve. At 4 steps the
$K{=}2$--$4$ student beats the fixed 4-step student (6.32 \vs 6.46), reaches its final FID in half the iterations and
has higher precision against teacher samples (0.737 \vs 0.707). The $K{=}3$--$6$ student keeps improving from 3 to 8
steps (8.16 to 6.24). With one fake score per $K$, a $K{=}2$--$4$ student peaks at 6.59 after 2k iterations and then
falls apart (11.21 at 6k). With 50k samples and 4 steps, the $K{=}2$--$4$ student scores FID 3.95 and IS 310.5
(\cref{tab:c2i_full}), and the fixed 4-step student trained as long (6k iterations) scores 4.01 and 278.4. Random
steps also keep the image stable as the step count changes (\cref{fig:qual_steps}).
\label{sec:supp_L}
On GRN-L the $K{=}3$--$6$ student keeps FID-10k between 5.49 and 5.66 from 4 to 8 steps, while the fixed 4-step
student climbs from 5.43 to 6.46.

\paragraph{Oversampling.}
The oversampling judge acts fast (\cref{tab:sb_full}). After 1k iterations the 2-step student is at 6.25, against 6.50
with the original judge. In effect, the round moves the third step into the first two. The starting student needed
three steps to beat the teacher (6.04 against 6.24), and after the round two steps reach its level. With 50k samples
the student reaches FID 3.83 and IS 306.8, against 4.12 and 280.4 with the original judge. The judge itself barely
moves. Its validation CE on held-out student samples falls from 1.5755 to 1.5733 nats per token, and a judge
fine-tuned the same way on teacher data (1.5870 $\to$ 1.5871) does no better than the unchanged one. What helps is a
judge that has seen the student's own refined samples (\cref{fig:qual_sb}).

\paragraph{Combination.}
Random steps leave refinement headroom, and oversampling turns it into fewer steps (\cref{fig:abl_os4}). On the
$K{=}3$--$6$ student the oversampling judge gives FID-10k 9.89 with 2 steps and 6.69 with 3, against 13.3 and 7.76
with the original judge, and 6.32 against 6.52 with 4 steps, level with the fixed 4-step student (6.33). With 50k
samples it gives FID 4.02 and IS 301.3 (\cref{tab:c2i_full}). On GRN-L, a round on the $K{=}3$--$6$ student gives 10.4
with 2 steps, against 12.8 with the original judge and 16.8 for the fixed 4-step student trained as long.

\begin{table}[t]
\centering
\caption{\textbf{The judge is the ceiling.} Per-token KL (nats) between posteriors at corrupted student samples, by
keep probability $q$ (6k-iteration 2-step GRN-B student, 256 samples). Notes: Sec.~\ref{sec:supp_tabnotes_s}.}
\label{tab:selffake}
\footnotesize
\setlength{\tabcolsep}{1.8pt}
\begin{tabular}{@{}lrrrrrrr@{}}
\toprule
$q$ & 0.05 & 0.1 & 0.2 & 0.3 & 0.4 & 0.6 & 0.8 \\
\midrule
$\KL(D_F\|D_R)$ & 0.0019 & 0.0013 & 0.0026 & 0.0032 & 0.0031 & 0.0030 & 0.0028 \\
$\KL(D_F\|D_\theta)$ & 0.673 & 0.689 & 0.711 & 0.631 & 0.551 & 0.398 & 0.245 \\
\bottomrule
\end{tabular}
\end{table}

\paragraph{Diagnostics.}
\label{sec:supp_diag}
After 6k iterations the 2-step student has caught up with its judge. The fake and real posteriors agree to
0.0013--0.0032 nats per token (\cref{tab:selffake}), so a better student needs a better judge
(Sec.~\ref{sec:sbr}). The student network cannot stand in for the fake score, because its prediction lies
0.24--0.71 nats per token away from it. Matching the first step as well, with its own fake score, gives a 2-step
FID-10k of 7.64--7.77 at 6k iterations against 6.54 for last-step matching, so the first step is best left a draft.

\subsection{Non-uniform sampling}
\label{sec:supp_eh}

\InputOrPH{tables/tab_eh_full.tex}{\TablePHwide{tab:eh_full}{Non-uniform sampling: full results}{6.0cm}}

\begin{figure*}[t]
  \centering
  \IncludeOrPH[width=\linewidth]{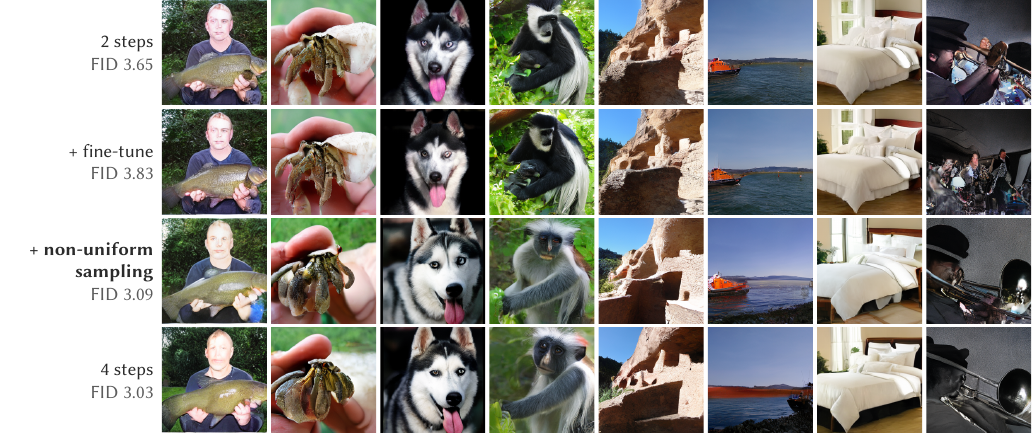}{\linewidth}{6.0cm}{\textbf{Non-uniform sampling samples placeholder}}
  \vspace{-1.4em}
  \caption{\textbf{At 2 NFE, the non-uniform-sampling student takes the composition of the 4-step student} (GRN-L,
  $256^2$, FID-50k; same class and noise per column). Middle rows: 2k more iterations on $\{0,0.5\}$ and
  $\{0,0.25\}$.}
  \label{fig:qual_eh}
\end{figure*}

\paragraph{GRN-L.}
Non-uniform sampling takes the base student from FID-50k 3.65 to 3.09 (\cref{tab:eh_full}, \cref{fig:qual_eh}) and
lifts its recall from 0.661 to 0.710. Most of the gain is learned. Sampled back on $\{0,0.5\}$, the student still
reaches 3.24. Every trained grid point beats the $\{0,0.5\}$ continuation (3.83), with a plateau at $0.25$ and $0.33$
(3.14) and $0.17$ (3.33) already past it. One student can even hold two budgets. Trained with 2 steps on
$\{0,0.25\}$ and 4 steps on the uniform grid, it gives FID-10k 5.731 and 5.285, against 5.945 and 5.527 when trained
on uniform grids only.

\paragraph{Why the draft step.}
We splice the four sampling steps one at a time between the fixed 4-step GRN-L student and the weaker $K{=}2$--$4$
student. Step 0, the draft from pure noise, carries 77--83\% of the recall gap and 74--95\% of the FID gap. This step
gets no gradient when $K\ge2$, and non-uniform sampling shortens this part of the path. The teacher shows the
same effect. Compressing its first 21 steps, which run without CFG, into 2 costs $+3.14$ FID-10k at 60 NFE, while a
uniform 38-step schedule at the same NFE costs only $+0.43$. Learned samplers~\cite{watson2022ddss}, the EDM
discretization~\cite{karras2022edm} and trajectory-regularity schedules~\cite{chen2024gits} choose the time steps of
a trained model, while we fix the grid before distillation.

\subsection{Non-uniform sampling on GRN-B}
\label{sec:supp_eh_b}

\InputOrPH{tables/tab_eh_b.tex}{\TablePH{tab:eh_b}{Non-uniform sampling on GRN-B}{8.0cm}}
\InputOrPH{tables/tab_eh_probe.tex}{\TablePH{tab:eh_probe}{What the two steps do}{4.0cm}}

Two thousand iterations on $\{0,0.25\}$ take the 2-step \tagW+\tagSB{} student from FID-50k 3.83 to 3.73
(\cref{tab:eh_b}a), while the same fine-tune on $\{0,0.5\}$ ends at 3.86. Sampled on the standard grid, the
$\{0,0.25\}$ student reaches 3.64, below the teacher's 3.73, so the gain is learned on the non-uniform grid and
carries back to the standard one. Trained on $\{0,0.33\}$, it reaches 3.71, a plateau as on GRN-L. Over six trained
grid points $t_1$ (\cref{tab:eh_b}b), precision rises monotonically from 0.673 to 0.704 and recall falls from 0.740 to
0.716. The \tagW{} student lacks precision (0.687 against the teacher's 0.733) more than recall (0.722 against 0.730),
so its best FID-50k comes at $t_1=0.6$.

\paragraph{What the two steps do.}
On 64 noises for each of 8 classes (\cref{tab:eh_probe}), the draft is nearly the same for every noise of a class.
Its predicted distributions vary much less across noises than the final ones (0.08--0.17 \vs 0.40--0.71). The last
step keeps 67--87\% of the committed draft tokens, far above the 21--29\% agreement on positions it never saw, and
adds what is specific to the sample. The grid point sets how much of the image the trained step decides (81\%, 60\%
and 45\% of the tokens at $t_1=0.25,0.5,0.67$), so a smaller $t_1$ buys recall at the cost of precision.

\subsection{Refinement confidence keeping}
\label{sec:supp_conf}

\InputOrPH{tables/tab_conf.tex}{\TablePH{tab:conf}{Refinement confidence keeping}{8.0cm}}

\paragraph{Teacher.}
Besides the tokens GRN keeps at random with probability $t_{s+1}$, keeping holds on to every token whose top
predicted probability is at least $a$. The next step still re-predicts everything, so the rule adds no network call.
On the GRN-B teacher (\cref{tab:conf}a) it takes the 25-step FID-10k from 9.59 to 6.63, and 16 steps with keeping beat
25 steps without it (8.33, with 25 NFE against 39). Keeping whole tokens works best. Keeping confident bits comes close
(6.67), and freezing confident tokens for good helps less (8.13). The threshold keeps almost no token in the first
steps and most open tokens in the last ones (12\% of the open tokens over 25 steps). Random tokens kept in the same
share step by step match keeping with 25 steps (6.65 against 6.72 on a second machine) and trail it with 16 (8.67
against 8.35), while the same share spread evenly falls behind at both (7.00 and 10.78). Most of the gain thus comes
from when tokens are kept, and confidence adds to it when the steps are few. Letting confidence replace the random
choice, as in GRN's own ablation~\cite{han2026grn}, breaks sampling (146.39 with 25 steps, 91.37 against 6.17 with
50), because the clean tokens then follow the content instead of the random set GRN is trained on.

\paragraph{Students.}
At sampling time alone, keeping takes the 2-step \method-B from 3.73 to 3.71 (\cref{tab:conf}b,c). Trained with
keeping ($a=0.51$) in every rollout of its non-uniform-sampling fine-tune, \method-L improves from 3.09 to 3.03 at 2
NFE, the FID of our 4-step student, mostly through recall (0.709 to 0.716), and \method-B stays at 3.73. Without this
training the same threshold hurts (3.82 and 3.11). A 2-step draft gives the rule little to work with. Its confidence
hardly predicts which committed draft tokens survive (AUC 0.54), and only 0.3--0.4\% of the open draft tokens reach
probability 0.9. Confidence still picks better tokens than chance, since keeping the same share at random costs
0.04--0.05 FID.

\subsection{The window}
\label{sec:supp_window}

\begin{figure}[t]
  \centering
  \IncludeOrPH[width=\linewidth]{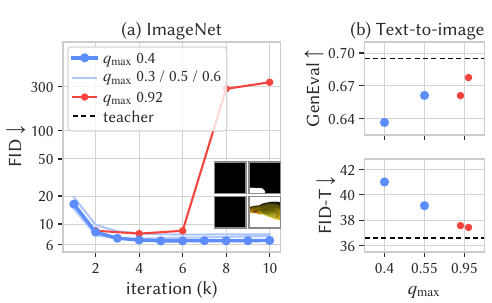}{\linewidth}{3.6cm}{\textbf{Window placeholder}}
  \vspace{-1.6em}
  \caption{\textbf{The window keeps ImageNet training stable, and text-to-image is more forgiving.} 2-step students; the window
  ends at $\qmax$ (0.92, 0.95: no window; two runs at 0.95). \textbf{(a)}~GRN-B, FID-10k; inset: no-window samples at
  8k. \textbf{(b)}~2k iterations; black dashed: teacher (re-sampled for FID-T).}
  \label{fig:window}
\end{figure}

\InputOrPH{tables/tab_ablation_c2i.tex}{\TablePH{tab:abl_c2i}{Ablations on ImageNet $256^2$}{3.4cm}}
\InputOrPH{tables/tab_c2i_fewstep.tex}{\TablePH{tab:fewstep}{Few-step sampling without distribution matching}{2.2cm}}

\paragraph{Why a window.}
Without the window ($\qmax=0.92$), the 2-step GRN-B student reaches FID-10k 7.87 at 4k iterations, collapses at 6.4k
(the fake score's per-token accuracy on student samples reaches 0.8) and gives mostly black or white images by 8k
(\cref{fig:window}a). Three forces push toward blank images. (i)~\emph{Attraction.} The cross-entropy part of the
reverse KL pulls mass toward high real density, and the fine-tuned GRN gives the collapsed images a NELBO of
0.05--0.12 nats per token, against 2.28 for teacher samples, as likelihood models favor simple
inputs~\cite{nalisnick2019know,serra2020input}. (ii)~\emph{A lagging fake score.} The entropy part that should push
back comes from the online fake score, which is least reliable near the clean end. (iii)~\emph{Absorption.} The factor
$\pi_b(\Delta_b-\E_\pi\Delta)$ in Eq.~(\ref{eq:grad}) vanishes as tokens saturate, and the rarity of blank images
shows only at $q\lesssim0.1$, where the gradient is scaled down by $q$. Cutting the near-clean end stops the collapse.
With $\qmax=0.4$ no run collapsed, including a 2-step run continued to 24k iterations.

\paragraph{Ablations.}
Without the window, matching collapses (FID-10k 332.8 at 10k), and every $\qmax$ from 0.3 to 0.6 stays stable
(\cref{fig:window}a). The value 0.6 slowly drifts, and 0.3 costs fidelity (7.42 and 7.75 \vs 6.65 at 0.4). The window
also helps the 4-step student, which does not collapse without it (FID-50k 4.63 $\to$ 3.99, \cref{tab:abl_c2i}b), and
a single level $q=0.4$ does as well (4.15 \vs 4.17). Bits are more forgiving (\cref{fig:window}b, \cref{tab:abl_t2i}c).
Without the window, the text-to-image student drifts in its training diagnostics while its images hold up (GenEval
0.676 at 8k). By per-token mutual information, 0.4 on bits corresponds to about 0.29 on ImageNet. Paired noise is the
better start (FID-10k 6.65 against 9.85 from the fine-tuned GRN, \cref{tab:abl_c2i}a, and in text-to-image FID-T 42.1
\vs 48.6 with 0.90 \vs 0.73 of the teacher's diversity), and the fine-tuned GRN is a better judge than the guided
teacher (FID-T 41.0 \vs 49.9, \cref{tab:abl_t2i}a,\,b).

\paragraph{Toy phase diagram.}
A $D=64$ Markov-chain toy with $V=8$ values, exact real posteriors, a learned fake score and a one-step generator
initialized from paired noise reproduces the collapse. The peak fraction of flat tokens (target about 0.02) is 0.61
at $\qmax=0.92$ and 0.24 at 0.7 (collapse), 0.093 at 0.6 (critical) and at most 0.039 at 0.3--0.5 (stable).
Re-opening the window of a healthy generator to 0.92 brings the collapse back (0.275), while opening it to 0.6 does
not (0.046). The unwindowed objective itself does not favor the collapsed state, yet training without the window
raised it (2.52 to 4.06), so the collapse comes from the biased gradient of the learned fake score near the clean end.

\subsection{Refinement and noise}
\label{sec:supp_refine}

\begin{figure}[t]
  \centering
  \IncludeOrPH[width=\linewidth]{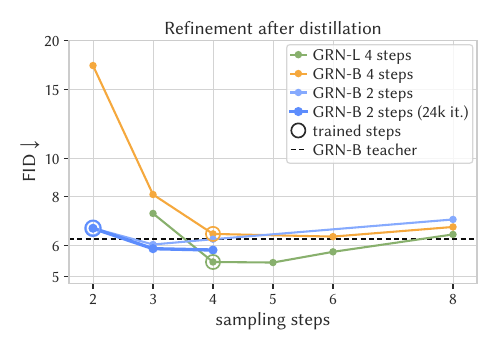}{\linewidth}{5.5cm}{\textbf{Refinement placeholder}}
  \vspace{-1.6em}
  \caption{\textbf{Refinement after distillation} (ImageNet $256^2$, FID-10k). Each curve is one trained \tagW{}
  student sampled with $N$ steps; rings mark its trained steps.}
  \label{fig:abl_refine}
\end{figure}
\begin{table}[t]
\centering
\caption{\textbf{Do the coefficients of Eq.~(\ref{eq:grad}) matter on toy problems?} $\KL(p\,\|\,p_\theta)$ (nats) of a trained
one-step generator; exhaustive computation, float64. Notes: Sec.~\ref{sec:supp_tabnotes_s}.}
\label{tab:toy}
\footnotesize
\setlength{\tabcolsep}{3.2pt}
\begin{tabular}{@{}lrrrrrr@{}}
\toprule
 & & & & \multicolumn{2}{c}{Oracle posteriors} & Learned \\
\cmidrule(lr){5-6}\cmidrule(l){7-7}
Target & $D$ & $V$ & $\TC(p)$ & Eq.~(\ref{eq:grad}) & Heuristic & fake \\
\midrule
Dirichlet & 4 & 3 & 1.045 & \textbf{0.0021} & 0.0856 & 0.0765 \\
Ising & 6 & 2 & 1.639 & \textbf{0.0001} & 0.0002 & 0.0247 \\
Prototypes & 4 & 3 & \na & \textbf{0.0000} & 0.0002 & 0.0191 \\
Parity & 4 & 2 & 0.495 & \textbf{0.0000} & 0.4947 & 0.0186 \\
\bottomrule
\end{tabular}
\end{table}

\begin{figure*}[t]
  \centering
  \IncludeOrPH[width=\linewidth]{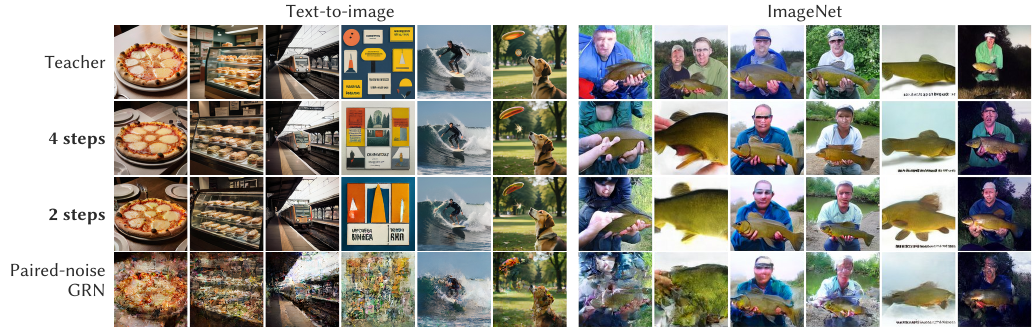}{\linewidth}{4.5cm}{\textbf{Qualitative samples placeholder}}
  \vspace{-1.6em}
  \caption{\textbf{At 4 and 2 steps \method gives clean samples, while its paired-noise initialization alone falls apart}
  (GRN-T2I 2B, GRN-B, $256^2$; same noise per column). \method{} rows: \tagW{} students; paired-noise GRN row: the initialization, 2 steps.}
  \label{fig:samples}
\end{figure*}
\InputOrPH{tables/tab_t2i_full.tex}{\TablePHwide{tab:t2i_full}{Text-to-image: all results}{3.0cm}}

\paragraph{Refinement after distillation.}
A distilled student is still a GRN, so it can be sampled with more steps than it was trained for
(\cref{fig:abl_refine}). The 4-step GRN-B student improves from FID-10k 6.42 to 6.32 with 6 steps, within 0.08 of its
79-NFE teacher. The 2-step student improves from 6.65 to 6.04 with 3 steps and, after 24k iterations, to 5.84 with 4
steps, below the teacher's 6.24. Too few steps hurt (17.3 for the 4-step student with 2 steps).

\paragraph{Frozen noise and paired noise.}
Paired-noise fine-tuning pays off because GRN's sampler freezes its noise. Redrawing the noise at every step raises the
8-step FID-10k of a paired-noise GRN-B from 13.8 to 101.6 and that of the fine-tuned GRN from 17.4 to 71.4, while the
distilled 2-step student cares much less (6.65 frozen \vs 7.37 fresh). Unlike Reflow~\cite{liu2023rectified}, paired
noise does not straighten trajectories. The samples of a paired-noise GRN-B change far less between 2 and 16 steps
than those of reference models (endpoint change 0.019 against 0.076 or more), yet its 16-step trajectories depend on
their noise no more than theirs. Paired noise mainly makes the noise informative.

\paragraph{Toy evidence for the per-token gradient.}
\label{sec:supp_toy}
On exhaustive toy problems (\cref{tab:toy}), the coefficients of Eq.~(\ref{eq:grad}) with oracle posteriors bring
the one-step KL to at most 0.0021 nats on all four targets. A posterior-difference heuristic is 40$\times$ worse on the
Dirichlet target and stalls at 0.4947 nats on the parity target, its total correlation, the product-of-marginals
failure of \cref{prop:tc}. With a learned fake score, the same coefficients leave 0.019--0.077 nats, so the fake score
is the bottleneck.

\subsection{Comparisons, text-to-image and cost}
\label{sec:supp_more}
\label{sec:supp_qual}
\InputOrPH{tables/tab_c2i_full.tex}{\TablePH{tab:c2i_full}{Class-conditional ImageNet, all rows and columns}{5.0cm}}
\InputOrPH{tables/tab_compare_ctx.tex}{\TablePH{tab:compare_ctx}{Native and continuous few-step generators}{4.0cm}}
\InputOrPH{tables/tab_t2i_ctx.tex}{\TablePH{tab:t2i_ctx}{Few-step text-to-image students}{4.0cm}}

\begin{figure}[t]
  \centering
  \IncludeOrPH[width=\linewidth]{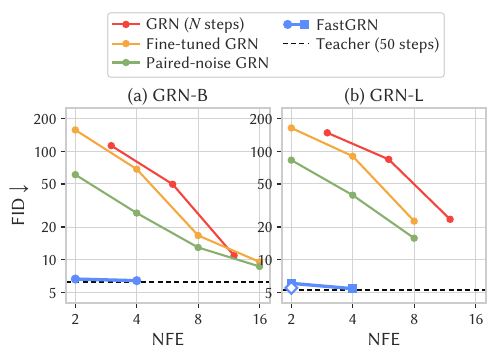}{\linewidth}{3.6cm}{\textbf{Quality vs.\ cost placeholder}}
  \vspace{-1.6em}
  \caption{\textbf{Few-step sampling without distribution matching fails} (ImageNet $256^2$, FID-10k). GRN with $N$ steps,
  the fine-tuned and the paired-noise GRN (fine-tuned on teacher samples with independent and with paired noise) use GRN's sampler; black dashed: teachers (79 NFE). Hollow diamond: 2-step \method-L with non-uniform
  sampling (5.51; base student, same machine: 6.04).}
  \label{fig:nfe}
\end{figure}

\InputOrPH{tables/tab_ablation_t2i.tex}{\TablePH{tab:abl_t2i}{Ablations on text-to-image}{3.0cm}}
\InputOrPH{tables/tab_speed.tex}{\TablePH{tab:speed}{Sampling cost per image}{2.2cm}}
\begin{figure*}[t]
  \centering
  \IncludeOrPH[width=\linewidth]{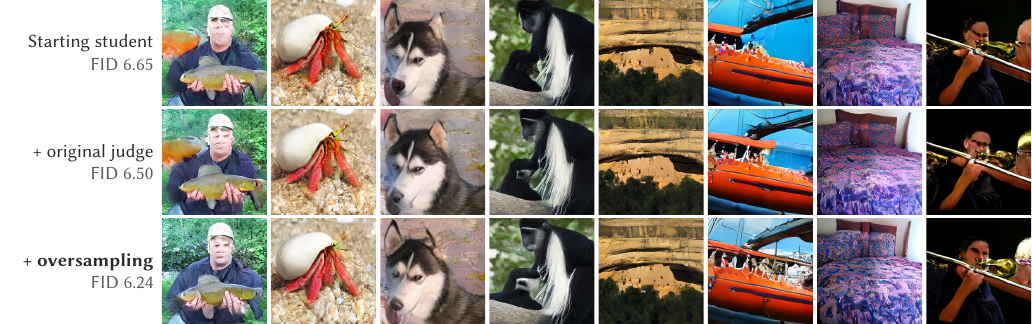}{\linewidth}{3.6cm}{\textbf{Oversampling samples placeholder}}
  \vspace{-1.6em}
  \caption{\textbf{The oversampling judge keeps each sample's composition, changing local detail} (GRN-B,
  2 steps, $256^2$, FID-10k; same class and noise per column). ``+'' rows: 3k more iterations with that judge.}
  \label{fig:qual_sb}
\end{figure*}

\begin{figure*}[t]
  \centering
  \IncludeOrPH[width=\linewidth]{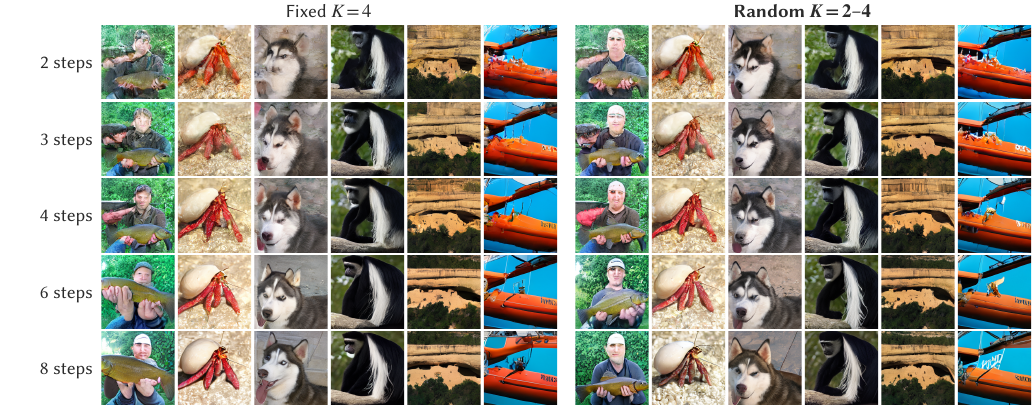}{\linewidth}{4.0cm}{\textbf{Step-count samples placeholder}}
  \vspace{-1.6em}
  \caption{\textbf{At 2 steps the fixed 4-step student smears textures and distorts shapes, while the random-step
  student stays coherent} (GRN-B, $256^2$; $K$: training steps; same class and noise per column).}
  \label{fig:qual_steps}
\end{figure*}

\paragraph{Class-conditional comparisons.}
Without distribution matching, few steps fail. Even at its best temperature, the teacher's own sampler scores FID-10k
49.6 with 4 steps, and the paired-noise GRN needs 16 steps to reach 8.7 (\cref{fig:nfe}, \cref{tab:fewstep}). GRN
itself reports no few-step results, and its step-reduction study reaches FID 3.79 with 24 adaptive steps on average.
Measured against each method's own teacher (Tab.~\ref{tab:compare}), the \tagW{} students land at
1.07--1.15$\times$ with 2--4 steps and the \tagW+\tagSB+\tagEH{} 2-step student at 1.00$\times$. With one step,
Di[M]O reaches 1.05$\times$. Soft-Di[M]O, trained with a GAN loss on real images, reaches 1.18$\times$ on MaskBit,
0.94$\times$ after longer training and 0.97$\times$ on MaskGIT. \Cref{tab:c2i_full} gives every row of
Tab.~\ref{tab:c2i} with precision, recall and latency, and \cref{tab:compare_ctx} adds more distilled students, the
native discrete samplers and few-step generators in continuous latent spaces. GRN-L with 4 steps trades a little
recall for precision (0.697 against 0.729 recall, 0.734 against 0.728 precision), as expected from the mode-seeking
reverse KL.

\paragraph{Text-to-image.}
\Cref{tab:t2i_full} lists every text-to-image result, \cref{tab:t2i_ctx} the few-step students of other models as
reported, \cref{tab:abl_t2i} the ablations and \cref{tab:speed} the cost numbers of Sec.~\ref{sec:c2i}.
\Cref{fig:samples} shows matched-noise samples. On 5k COCO-2014 validation captions, scored against all 40{,}504
validation images, the 2-step student ($\qmax=0.4$, 4k) reaches FID 26.2 and the CFG-4.5 teacher 29.6, and the
student's FID to the teacher's images is 12.4.

\begin{figure*}[tp]
  \centering
  \IncludeOrPH[width=\linewidth]{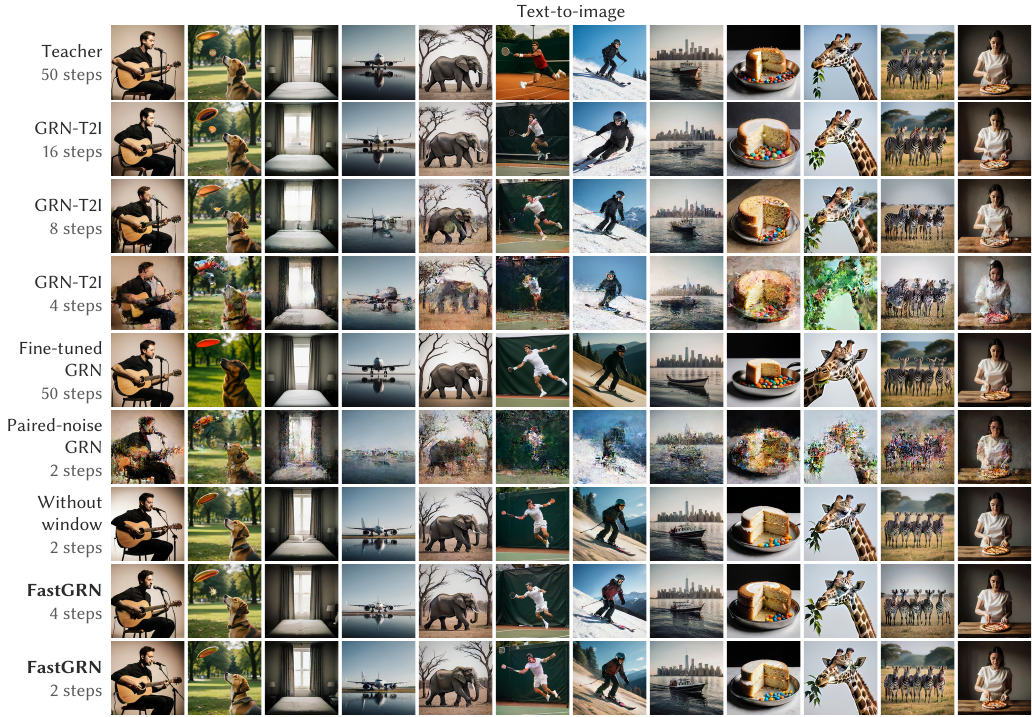}{\linewidth}{12.3cm}{\textbf{Visual comparison placeholder}}
  \vspace{-1.4em}
  \caption{\textbf{Text-to-image, same noise} (GRN-T2I 2B; prompts 1--12 of the fixed 24, one frozen noise per
  column). Fine-tuned GRN: fine-tuned on teacher samples, 50 steps without CFG; paired-noise GRN: fine-tuned on noise--sample pairs;
  without window: matching without the window.}
  \label{fig:compare_t2i_a}
\end{figure*}

\begin{figure*}[tp]
  \centering
  \IncludeOrPH[width=\linewidth]{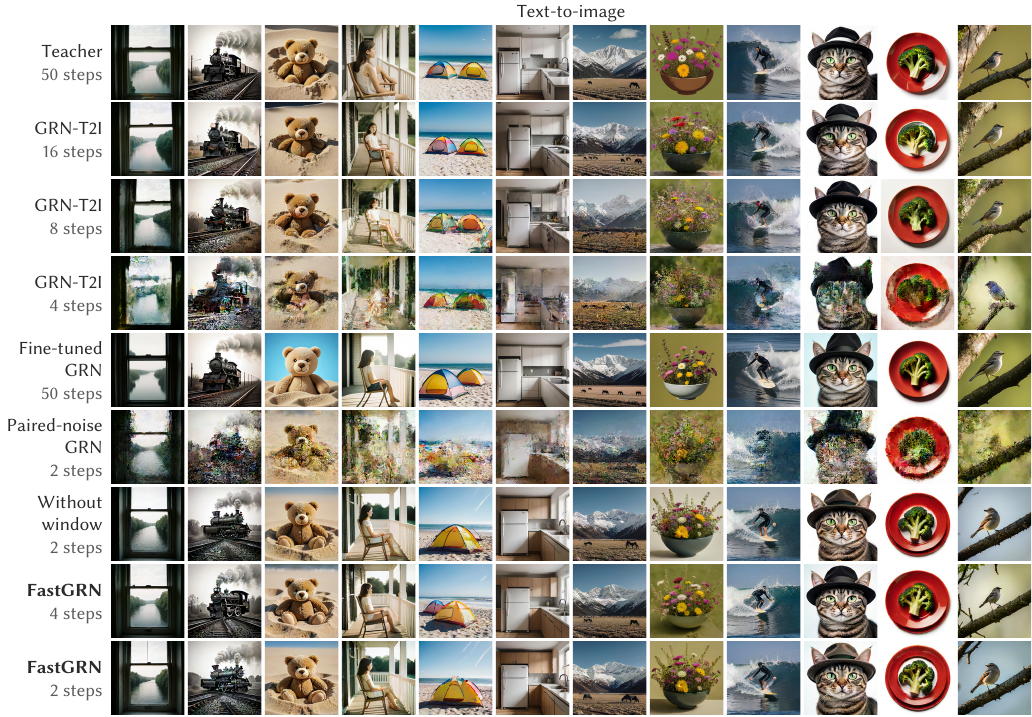}{\linewidth}{12.3cm}{\textbf{Visual comparison placeholder}}
  \vspace{-1.4em}
  \caption{\textbf{Text-to-image, same noise}, prompts 13--24.}
  \label{fig:compare_t2i_b}
\end{figure*}

\begin{figure*}[tp]
  \centering
  \IncludeOrPH[width=\linewidth]{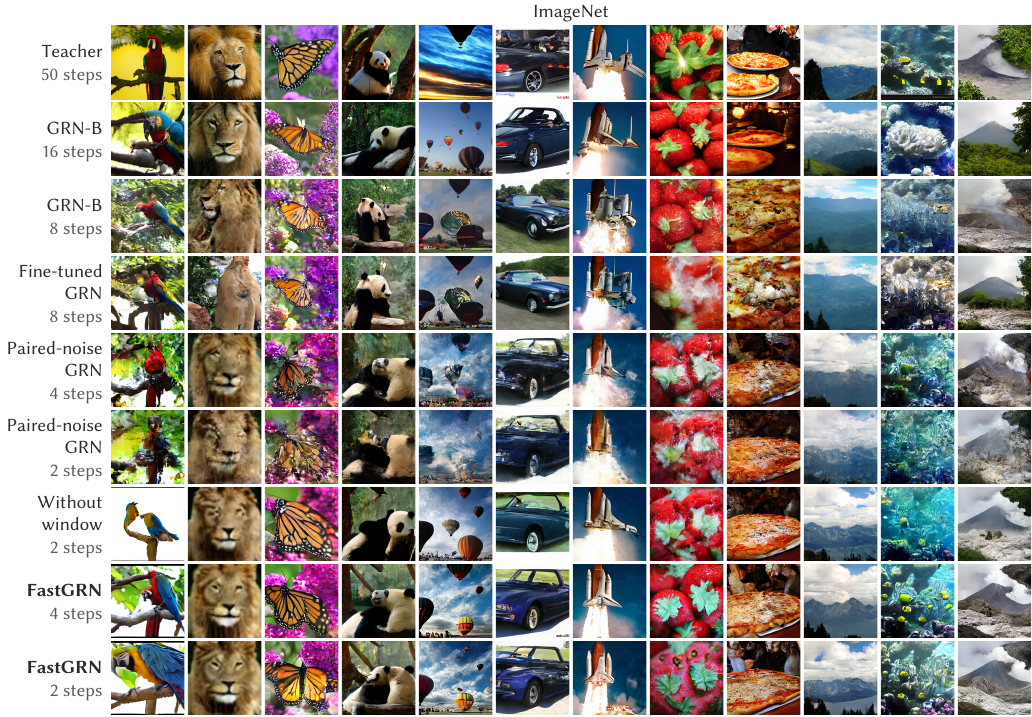}{\linewidth}{12.3cm}{\textbf{Visual comparison placeholder}}
  \vspace{-1.4em}
  \caption{\textbf{ImageNet, same noise} (GRN-B; second sample of each class from the 50k evaluations of
  Tab.~\ref{tab:c2i}). Fine-tuned and paired-noise GRN: GRN fine-tuned on teacher samples with independent and with paired noise;
  without window: matching without the window, best checkpoint before it collapses.}
  \label{fig:compare_c2i_b}
\end{figure*}

\begin{figure*}[tp]
  \centering
  \IncludeOrPH[width=\linewidth]{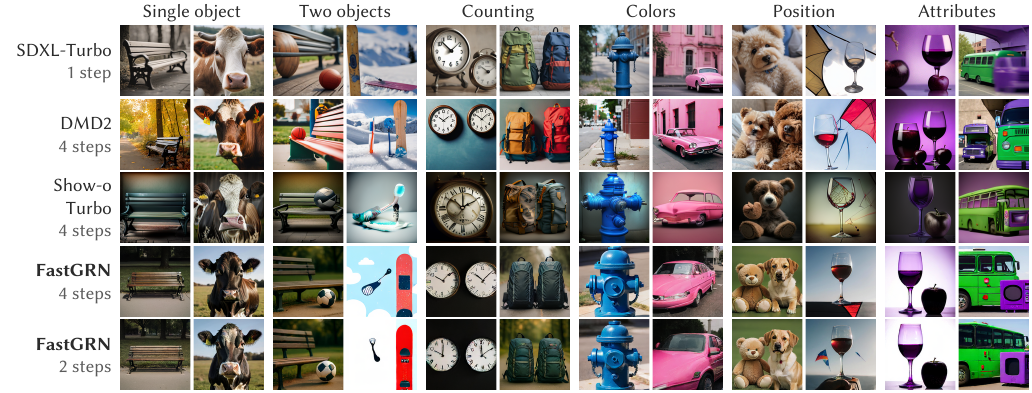}{\linewidth}{6.6cm}{\textbf{SOTA comparison placeholder}}
  \vspace{-1.4em}
  \caption{\textbf{Text-to-image against other few-step models} (GenEval prompts, the first two of each category; each
  model's released sampler; all shown at $256^2$).}
  \label{fig:compare_sota_t2i_a}
\end{figure*}

\begin{figure*}[tp]
  \centering
  \IncludeOrPH[width=\linewidth]{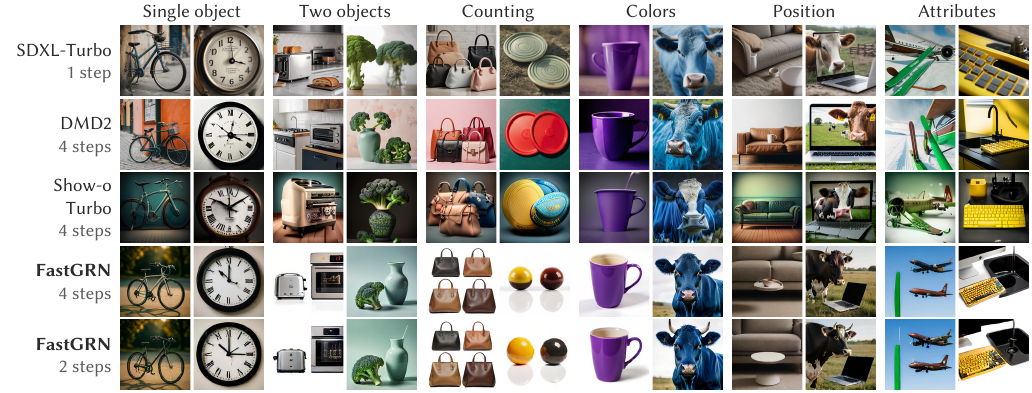}{\linewidth}{6.6cm}{\textbf{SOTA comparison placeholder}}
  \vspace{-1.4em}
  \caption{\textbf{Text-to-image against other few-step models}, the third and fourth prompt of each category.}
  \label{fig:compare_sota_t2i_b}
\end{figure*}

\begin{figure*}[tp]
  \centering
  \IncludeOrPH[width=\linewidth]{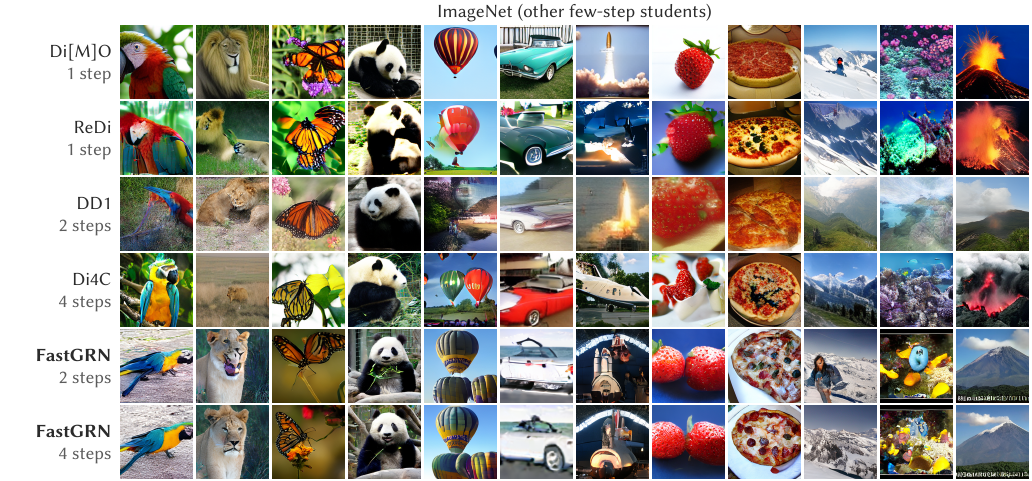}{\linewidth}{8.0cm}{\textbf{SOTA comparison placeholder}}
  \vspace{-1.4em}
  \caption{\textbf{ImageNet against other few-step students} (the GAN-free students of discrete generators in
  Tab.~\ref{tab:compare} with public weights; each model's released sampler and settings; one image per class).}
  \label{fig:compare_sota_c2i}
\end{figure*}

\paragraph{Visual comparisons.}
Figs.~\ref{fig:compare_t2i_a}--\ref{fig:compare_c2i_b} show every few-step route of the same teacher for 24
text-to-image prompts and 12 ImageNet classes, with one frozen noise per column and the prompts and classes fixed
before any image was generated. Figs.~\ref{fig:compare_sota_t2i_a} and~\ref{fig:compare_sota_t2i_b} extend
Fig.~\ref{fig:compare} to 24 GenEval prompts, the first four of each category. \method{} draws the requested number
of objects and their relations more often, for example two clocks, a dog right of a teddy bear and a yellow keyboard
next to a black sink. The other students tend to merge two objects into one, such as a cow inside a laptop screen, or
to drop one of them. Every model misses some prompts, and \method{} draws six handbags for four. On ImageNet
(\cref{fig:compare_sota_c2i}), ReDi with 1 step blurs textures, DD1 with 2 steps often loses the object, such as the
balloon and the coral reef, Di[M]O draws clean, centred objects on smooth backgrounds, and Di4C with 4 guided steps
distorts some of them, such as the lion. \method keeps the photographic texture and the scene layout with 2 steps. In
\cref{fig:qual_sb,fig:qual_steps} each column fixes the class and the frozen noise (tench, hermit crab, Siberian
husky, black-and-white colobus, cliff dwelling, lifeboat, quilt and trombone).

\section{Notes to the supplementary tables}
\label{sec:supp_tabnotes_s}
\emph{Table~\ref{tab:sb_full}.} Top: 3k-iteration continuations of the 10k-iteration 2-step \tagW{} student, with the fake score initialized from the judge; the last oversampling row gives the FID$_\mathrm{50k}$ of Tab.~\ref{tab:c2i}. Original judge: GRN fine-tuned on teacher samples (EMA 0.9995); oversampling judge $R'$: the original judge fine-tuned on 200k of the student's own $N{=}3$ samples; teacher-data judge $R_T$: the same recipe on the teacher samples the original judge was trained on. P$_\mathrm{T}$\,/\,R$_\mathrm{T}$\,/\,FID-T against 50k teacher samples; \na: not run. Teacher: FID-10k 6.24, FID$_\mathrm{50k}$ 3.73, IS$_\mathrm{50k}$ 305.9. Bottom: validation cross-entropy (nats per token, held-out samples of the same source) before and after the judge fine-tune.

\emph{Table~\ref{tab:selffake}.} $D_F$: online fake score; $D_R$: real score; $D_\theta$: the student network's own prediction at $(\vz',q)$. Live student weights, $\tau=0.8$.

\emph{Table~\ref{tab:eh_full}.} Continuations of the 10k-iteration \tagW{} 2-step GRN-L student, the base student (2000 iterations, \cref{sec:supp_eh_impl}), trained and sampled on the grid of the row. P$_\mathrm{T}$, R$_\mathrm{T}$, FID-T against 50k GRN-L teacher samples ($k{=}3$). (b): grid point $t_1$ of training and sampling, against the $\{0,0.5\}$ fine-tune; $^\ast$against the untrained base student on $\{0,0.5\}$ (6.066), +0.540 at 50k; the student with $K{=}2$ and $4$ is trained on $\{0,0.25\}$ for 2 steps and on the uniform grid for 4 steps. Teacher: FID$_\mathrm{50k}$ 2.65.

\emph{Table~\ref{tab:eh_b}.} Each student continues a 2-step student of Tab.~\ref{tab:c2i} for 2000 iterations, trained and sampled on the grid of the row. P$_\mathrm{T}$, R$_\mathrm{T}$ ($k{=}3$) against 50k GRN-B teacher samples. Teacher: FID$_\mathrm{50k}$ 3.73, P$_\mathrm{T}$ 0.733, R$_\mathrm{T}$ 0.730. Bold: lowest FID$_\mathrm{50k}$ in (a) and in the trained rows of (b).

\emph{Table~\ref{tab:eh_probe}.} Diversity: mean over noise pairs of half the squared distance between the predicted token distributions of two noises of one class, for the draft (first step) and the final prediction. Kept: share of the committed draft tokens that the last step keeps; Chance: the same agreement on positions that were not committed. Last step: share of the tokens the trained last step decides, $1-t_1\times$Kept. +2k rows: trained and sampled on $\{0,t_1\}$.

\emph{Table~\ref{tab:conf}.} Freezing: confident tokens stay fixed for all later steps. Bit rules: the same for the confident bit prefix of each token. Top-confidence: the most confident tokens replace the random choice (GRN's own ablation). Random: tokens kept at random in keeping's extra share, per step or in total. Extra: share of the open draft tokens that the rule keeps as well. Second block and 50-step column: a second L40S (native 25 steps: 9.59). Teacher: GRN's CFG and temperature (1.33) with 25, 16 and 50 steps (39, 25 and 79 NFE), $a{=}0.55$ and 0.9 per bit for the bit rules. Trained with keeping: the non-uniform-sampling fine-tune of Tab.~\ref{tab:c2i} repeated with keeping at $a{=}0.51$ in every rollout. P$_\mathrm{T}$, R$_\mathrm{T}$ ($k{=}3$) against 50k samples of each student's teacher. Bold: lowest FID per column and block, and per panel in (b), (c).

\emph{Table~\ref{tab:abl_c2i}.} \tagW{} students and their no-window and initialization variants; EMA generators; teacher FID-10k 6.24. ``Best'': best 1k-spaced milestone before a collapse. Collapse: first iteration at which the fake model's per-token accuracy on student samples reaches 0.8 (healthy runs: about 0.45). No window: $q\in[0.02,0.92]$; bold: \regrn{} (paired-noise init., $q_{\max}{=}0.4$). Anchor: extra loss toward the teacher's paired samples.

\emph{Table~\ref{tab:fewstep}.} One network call per step, except $^n$the teacher's own sampler with CFG: 3/6/12 NFE for 2/4/8 steps. Temperatures: on GRN-B the one with the lowest FID-10k per cell (with GRN's 1.33 the teacher's own sampler scores 63.5 at 8 steps); on GRN-L, 1.0 for the teacher's own sampler and 0.8 for the fine-tuned models. $^c$Collapses later in training. $^h$Non-uniform sampling on $\{0,0.25\}$; its base student scores 6.04 on the same machine.

\emph{Table~\ref{tab:toy}.} $D$ tokens with $V$ values each. The generator is a free table of per-token logits for each of $V^D$ latent values, which can represent every target exactly. 12 corruption levels in $[0.05,0.95]$. Heuristic: the logit gradient $D_F(\cdot\mid\vz)-D_R(\cdot\mid\vz)$, in the spirit of Di[M]O~\cite{zhu2025dimo}. Learned fake: the generator-side posterior is a model trained online.

\emph{Table~\ref{tab:t2i_full}.} $\Delta$ vs.\ T.: GenEval difference to the teacher. FID-T: FID to 1k teacher images for the same captions; $^a$the teacher against a second teacher sample set. CLIP: CLIP-L/14 $\times 100$. HPS: HPS~v2.1. Div.: same-prompt LPIPS diversity relative to the teacher. $q_{\max}$ 0.55$\to$0.95: the 2-step student with 0.55 continued for 1k iterations at 0.95 (CLIP and HPS not measured). $^\ast$Reported by GRN for its $1024^2$ checkpoint with a prompt rewriter.

\emph{Table~\ref{tab:c2i_full}.} GRN-B with 8 to 16 steps: the teacher's own sampler at the temperature with the lowest FID-10k (0.7, 0.8, 0.8, 1.0 and 1.0 for 8, 10, 12, 14 and 16 steps; CFG on $t\in(0.4,1)$ adds network calls). Fine-tuned and paired-noise GRN: fine-tuned on teacher samples with independent and with paired noise. Matching without window: its best checkpoint (4k iterations) before it collapses. Latency on one RTX 4090; the paired-noise GRN and matching without window share the latency of the students with the same steps, and the other baseline rows are not timed. Tags: \tagW{} paired-noise initialization and windowed matching (10k iterations); \tagRS{} random steps over $K{=}2$--$4$ (6k); \tagSB{} one oversampling round (+3k); \tagW+\tagRS+\tagSB{}: a $K{=}3$--$6$ student plus one round; \tagEH{} non-uniform sampling, +2k iterations on $\{0,0.25\}$; \tagCK{} refinement confidence keeping in that fine-tune and when sampling. Rows with more steps than trained: the same student sampled with 3 steps on $\{0,1/3,2/3\}$ (GRN-B) or $\{0,0.25,0.5\}$ (GRN-L), or with 4 steps on $\{0,0.23,0.54,0.68\}$, without training (3-step latency not measured). GRN-L \tagW+\tagEH+\tagCK+\tagSB: one round on the 2-step \tagW+\tagEH+\tagCK{} student with the judge fine-tuned on its own 3-step samples (+2k). The first two GRN-L 3-step rows: that student after 2k more iterations with a learning rate decaying to 5\% over the last 1k, without and with keeping when sampling; the last: the student itself. All students use EMA 0.999. Bold: best student FID.

\emph{Table~\ref{tab:compare_ctx}.} $^\dagger$Size not stated in the paper; the MaskGIT it starts from (Tab.~\ref{tab:compare}). $^c$Sampling steps as reported, with classifier-free guidance at inference. Teacher FID of the DD1 and DD2 rows as each paper reports it: VAR-d16 4.19 and 4.15, VAR-d20 3.35 and 3.40, VAR-d24 2.51 and 2.86, LlamaGen-B 5.42, LlamaGen-L 4.11; Di4C: its 8-step MaskGIT (6.57). SDTT and ReDi: as run by ReDi, whose MaskGIT has no 16-step reference. MaskGIT-PyTorch: the teacher of the MaskGIT students with its own sampler, as run by Di[M]O. MaskGIT and TiTok: their own 8-step samplers without guidance, no distillation. Token-Critic: 18 generator and 18 critic calls. DDPD: a planner and a denoiser, two calls per step. ReCAP: 8 full and 8 cached calls. VAR-d16 with its first scales skipped: as run by DD1. MeanFlow, $\alpha$-Flow, iMF, Drifting and SoFlow are trained from scratch in the SD-VAE latent space; DMF turns a flow model into a flow map, and FreeFlow distills a SiT-XL/2 flow model (teacher FID 2.06). They fold guidance into training, so one step is one network call; IMM uses CFG at its one step (two calls), and Shortcut's step count is as reported.

\emph{Table~\ref{tab:t2i_ctx}.} Each model at its own resolution; Show-o Turbo without CFG at $256^2$ and $512^2$ (2, 4, 8 steps). SANA-Sprint and FLUX.1-schnell as reported by SANA-Sprint (its Tab.~2, $1024^2$), whose SANA teachers score 0.64 (0.6B) and 0.66 (1.6B); the other continuous rows are taken from Soft-Di[M]O's Tab.~3, which evaluates all of them but InstaFlow from public checkpoints. $^g$GAN loss; $^r$reward fine-tuning; $^s$TTS: test-time search guided by reward models. Bold: highest among the students of discrete generators. \regrn: GRN-T2I 2B at $256^2$, 4 and 2 steps (Fig.~\ref{fig:t2i}).

\emph{Table~\ref{tab:abl_t2i}.} GenEval: 553 prompts, 4 images each; FID-T and HPS on the 1k COCO captions; Div.: LPIPS diversity relative to the teacher. (a): no GenEval at 1k. Parentheses in (c): paired GenEval difference to $q_{\max}{=}0.4$. Healthy: in (a, b), the training-side health check (diversity, same-prompt agreement, entropy, collapse guard) at 2k; in (c), no collapse signal up to the row's iteration. $^v$Training-side warnings only (collapse-guard votes at 5.5--5.55k, 7.4--7.5k, 7.65--8k iterations); the images stay intact (FID-T 37.4--38.4 at 1k--8k). $^w$vs.\ $q_{\max}{=}0.4$ at 4k, the longest windowed run. $^\ast$, $^\dagger$: two training runs without the window; $^\dagger$ is trained to 8k.

\emph{Table~\ref{tab:speed}.} Throughput: generation only on an A100-40GB (C2I, batch 250--500, \texttt{torch.compile}) or on an RTX 4090 ($^s$T2I, batch 64). ``Net'' is the generation speed-up over the teacher; ``+Dec.'' also counts the HBQ decoder (C2I: about 20\,ms per image; T2I: about 25\,ms for the students, 37\,ms in the teacher run). Batch 1: median end-to-end latency including decoding (ms) and the gain over the teacher.

\end{document}